\documentclass{article}
\usepackage{arxiv}

\usepackage{mathtools} 

\usepackage{booktabs}
\usepackage{array,multirow}
\usepackage{rotating}
\usepackage{makecell}
\usepackage{rotating}
\usepackage{array}
\usepackage{float}
\usepackage{physics}
\usepackage{algpseudocode}
\usepackage{enumitem}
\usepackage[font={small,it}]{caption}
\usepackage{hyperref}
\usepackage{microtype}

\usepackage{amssymb}
\usepackage{amsmath}
\usepackage{mathdots}
\usepackage{mathtools}
\usepackage{multicol}

\usepackage{tensor}

\usepackage{urwchancal}
\DeclareFontFamily{OT1}{pzc}{}
\DeclareFontShape{OT1}{pzc}{m}{it}{<-> s * [1.10] pzcmi7t}{}
\DeclareMathAlphabet{\mathpzc}{OT1}{pzc}{m}{it}

\usepackage{graphicx}
\usepackage{ifpdf}

\ifpdf
\usepackage{epstopdf}
\PrependGraphicsExtensions{.svg}
\fi

\usepackage{booktabs}
\usepackage{array,multirow}
\usepackage{rotating}
\usepackage{makecell}
\usepackage{rotating}
\usepackage{array}
\usepackage{float}
\usepackage{physics}
\usepackage{algpseudocode}
\usepackage{bbding}
\usepackage{amsmath}
\usepackage{bm}

\usepackage{wrapfig}

\usepackage{adjustbox}
\usepackage{multirow}
\usepackage[dvipsnames]{xcolor}
\usepackage[normalem]{ulem}
\usepackage{tikz}
\usepackage[ruled, vlined]{algorithm2e}
\makeatletter
\newcommand{\removelatexerror}{\let\@latex@error\@gobble}
\makeatother
\usepackage{array}
\usepackage{booktabs}
\usepackage{mathrsfs}

\usepackage{xcolor}
\usetikzlibrary{shapes, arrows}
\tikzset{%
	block/.style = {draw, thick, rectangle, minimum height = 2em, minimum width = 2em, node distance = 2cm},
	sum/.style = {draw, thick, circle, node distance = 2cm}, 
	gain/.style = {draw, thick, regular polygon, regular polygon sides=3,
		draw, fill=white, text width=1em,
		inner sep=0mm, outer sep=0mm,
		shape border rotate=-90, node distance = 1.6cm},
	invgain/.style = {draw, thick, regular polygon, regular polygon sides=3,
		draw, fill=white, text width=1em,
		inner sep=0mm, outer sep=0mm,
		shape border rotate=90, node distance = 1.6cm},
	input/.style = {coordinate}, 
	output/.style = {coordinate} 
}
\newcommand{\suma}{\small$+$}

\newcommand{\publicationdetails}
{Wang, Z., Yuan, D., Ng, Y., Mahony R. (2022), ``A Linear Comb Filter for Event Flicker Removal", \textit{published in IEEE International Conference on Robotics and Automation (ICRA), 2022.}
\copyrightNoticeIEEE{2022}	\\
	}
\newcommand{\publicationversion}
{Author Accepted Version}

\title{\LARGE \bf
	A Linear Comb Filter for Event Flicker Removal}

\author{
	\href{https://orcid.org/0000-0003-0815-1287}{\includegraphics[scale=0.06]{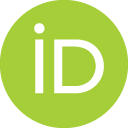}\hspace{1mm}
		Ziwei Wang}
	\\
	Systems Theory and Robotics Group \\
	Australian National University \\
	ACT, 2601, Australia \\
	\texttt{ziwei.wang1@anu.edu.au} \\
	\\
	\And
	\href{https://orcid.org/0000-0002-0412-4735}{\includegraphics[scale=0.06]{orcid.png}\hspace{1mm}
		Dingran Yuan}
	\\
	Systems Theory and Robotics Group \\
	Australian National University \\
	ACT, 2601, Australia \\
	\texttt{dingran.yuan@anu.edu.au} \\
	\And
	\href{https://orcid.org/0000-0002-7764-298X}{\includegraphics[scale=0.06]{orcid.png}\hspace{1mm}
		Yonhon Ng}
	\\
	Systems Theory and Robotics Group \\
	Australian National University \\
	ACT, 2601, Australia \\
	\texttt{yonhon.ng@anu.edu.au} \\
	\\
	\And	\href{https://orcid.org/0000-0002-7803-2868}{\includegraphics[scale=0.06]{orcid.png}\hspace{1mm}
		Robert Mahony}
	\\
	Systems Theory and Robotics Group \\
	Australian National University \\
	ACT, 2601, Australia \\
	\texttt{robert.mahony@anu.edu.au} \\
}

\begin{document}

\maketitle
\thispagestyle{empty}
\pagestyle{empty}


\begin{abstract}
	Event cameras are bio-inspired sensors that capture per-pixel asynchronous intensity change rather than the synchronous absolute intensity frames captured by a classical camera sensor.
	Such cameras are ideal for robotics applications since they have high temporal resolution, high dynamic range and low latency.
	However, due to their high temporal resolution, event cameras are particularly sensitive to flicker such as from fluorescent or LED lights.
	During every cycle from bright to dark, pixels that image a flickering light source generate many events that provide little or no useful information for a robot, swamping the useful data in the scene.
	In this paper, we propose a novel linear filter to preprocess event data to remove unwanted flicker events from an event stream.
	The proposed algorithm achieves over 4.6 times relative improvement in the signal-to-noise ratio when compared to the raw event stream due to the effective removal of flicker from fluorescent lighting.
	Thus, it is ideally suited to robotics applications that operate in indoor settings or scenes illuminated by flickering light sources.
\end{abstract}

\centerline{
	\noindent \textbf{Code, Datasets and Video:}
}
\centerline{
	\noindent \href{https://github.com/ziweiWWANG/EFR}{{\color{pink}\texttt{https://github.com/ziweiWWANG/EFR}}}
}

\section{INTRODUCTION}
Event cameras, such as the Dynamic Vision Sensor (DVS)~\cite{lichtsteiner2008128}, are bio-inspired vision sensors that record per-pixel brightness changes.
In contrast to conventional cameras that accumulate average brightness for every pixel over the exposure time to generate synchronous intensity frames, event cameras output a sparse asynchronous stream of positive and negative event data only when the change in logarithmic brightness at individual pixels exceeds a contrast threshold.
The high dynamic range (HDR), fine temporal resolution, low latency (\textless 0.5 ms) and sparse output~\cite{gallego2019event} make event cameras ideal for high-speed robotics vision applications that requires low computational cost and operate in challenging lighting conditions, such as feature tracking~\cite{Tedaldi16ebccsp,Liu16iscas, Gehrig18eccv}
de-blurring~\cite{Pan20pami,Jiang20cvpr}, motion estimation~\cite{kim2021real},
autonomous-driving applications~\cite{Maqueda18cvpr}, visual odometry and simultaneous localization and mapping (SLAM)~\cite{Kueng16iros,Rosinol18ral,liu2021spatiotemporal}.
However, these same properties make event cameras highly susceptible to flicker, particularly those caused by fluorescent and LED light sources, which can swamp the signal of interest in the scene.

\begin{figure}[t!]
	\centering
	\begin{tabular}{ c c }
		\\	\includegraphics[width=0.33\linewidth]{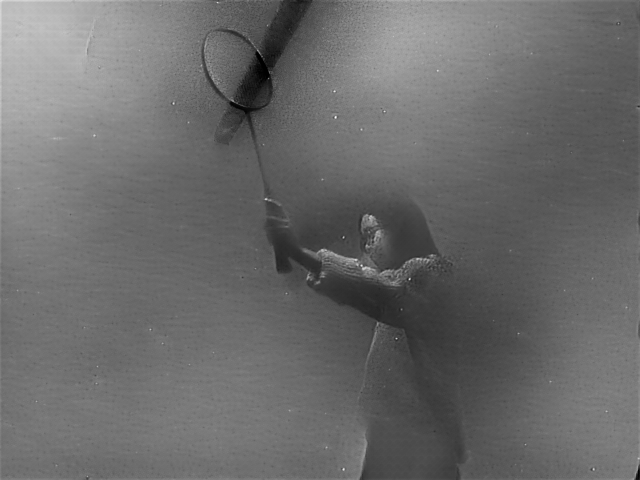} &
		\includegraphics[width=0.33\linewidth]{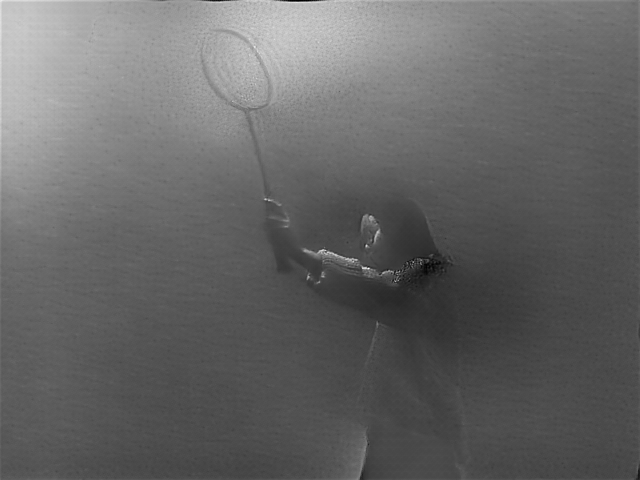} \\
		(a) & (b) \\ \hspace{5mm}
		\includegraphics[width=6.5cm]{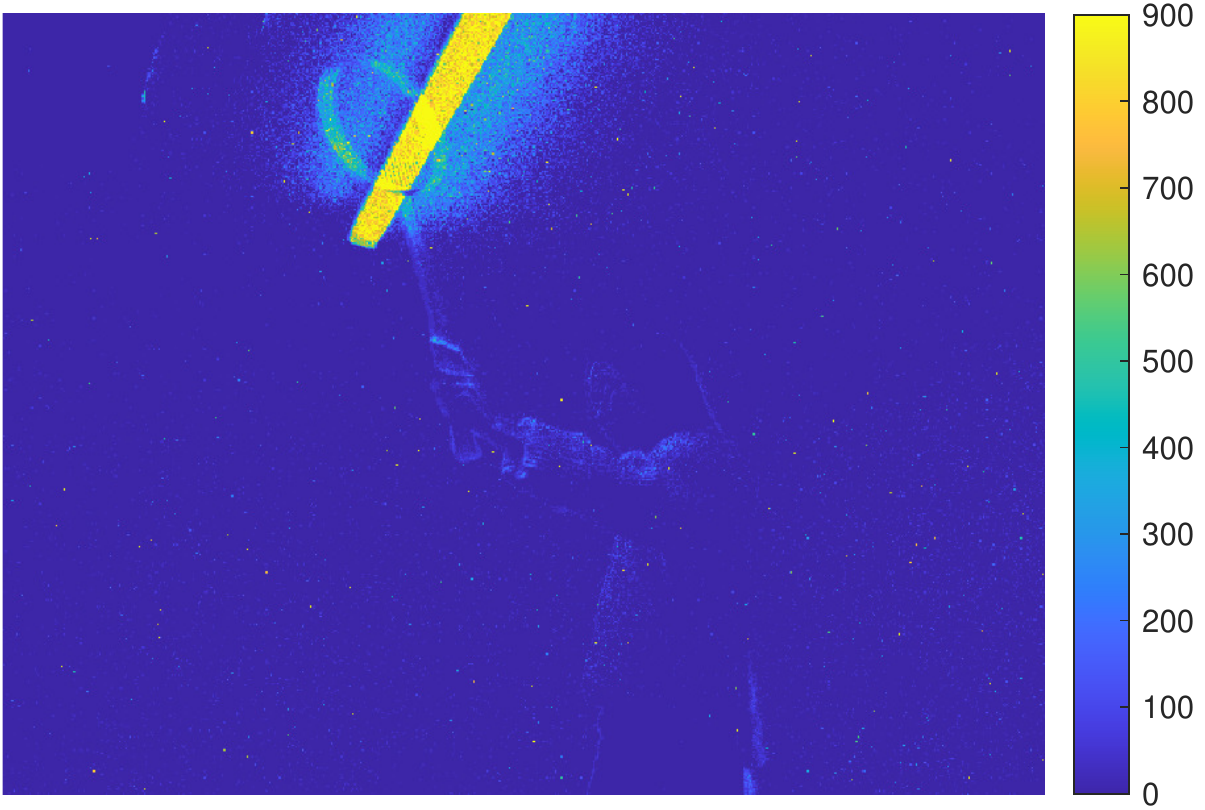} &
		\includegraphics[width=6.5cm]{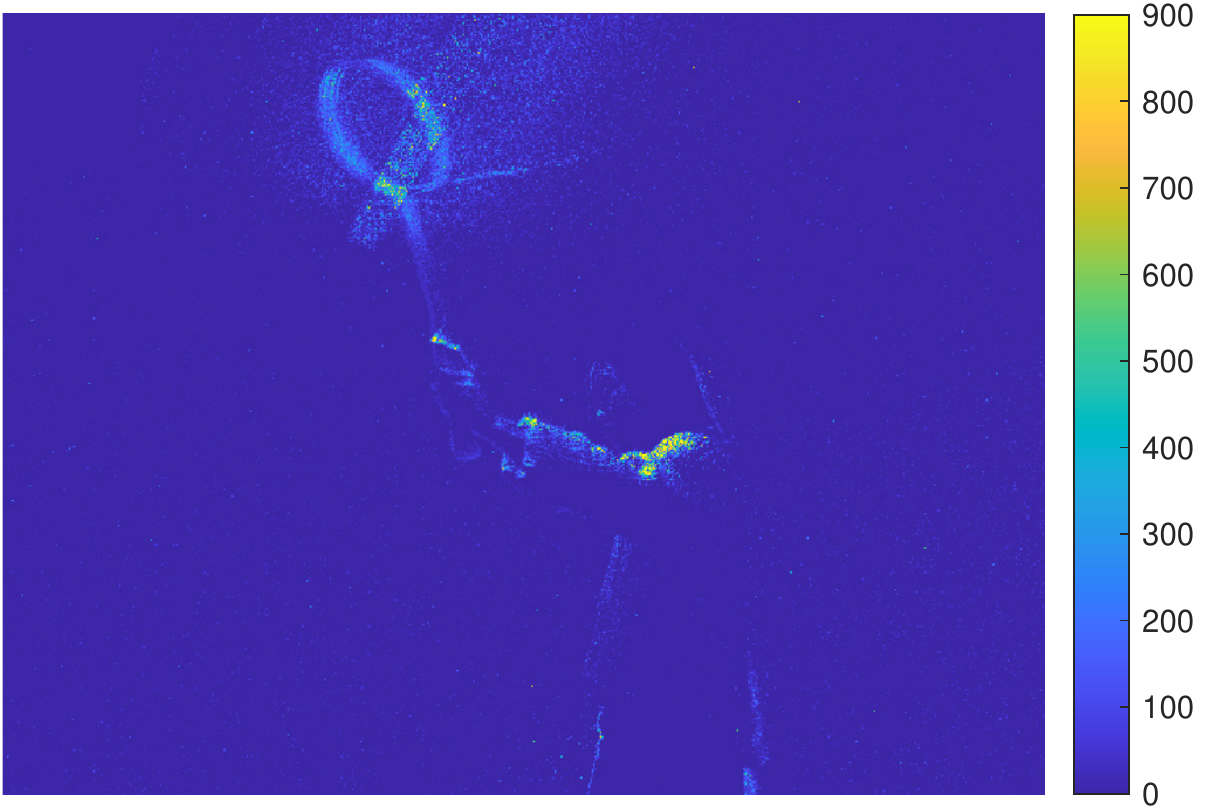} \\ \hspace{-2mm}
		(c) & \hspace{1mm} (d)  \\
	\end{tabular}
	\caption{
		(a) Image reconstruction E2VID~\cite{Rebecq20pami} from a high-speed dataset with a flickering (fluorescent) light source in the background.
		(b) Image reconstruction E2VID~\cite{Rebecq20pami} after event flicker removal using the proposed algorithm.
		(c) The heat map of events per second from the unfiltered event stream.
		(d) The heat map of events per second after event flicker removal.
	}\label{fig:front page}
\end{figure}

Event data de-noising is common in all current event cameras.
Current de-noising methods focus primarily on uncorrelated background activity noise~\cite{Delbruck08issle,Brandli14iscas,ieng2014asynchronous,liu2015design,khodamoradi2018n,padala2018noise,wang2019threshold}.
This noise is commonly caused by transistor switch leakage or thermal noise~\cite{Delbruck08issle,liu2015design} and is present even when there is no real brightness changes~\cite{liu2015design}.
Flicker removal is an entirely different problem since it corresponds to real intensity changes in the image and is only considered `noise' since flicker is uninteresting and distracting to most robotic systems.
However, in a typical indoor or nighttime environment with fluorescent or LED light sources, the events associated with the flicker of the lighting can overwhelm other events in the scene.
In the datasets collected in this paper, with only a single fluorescent light source, the unwanted events associated with the flickering light source outnumbered interesting foreground events and occupied 84\% of the event data on average.
This creates many problems in real-world robotics systems.
Excessive events due to flicker can overwhelm the onboard data bus leading to the lost events or requiring the camera to be set at lower sensitivity (leading to fewer events overall), resulting in a bigger quantisation error in the desired foreground information.
Modern image reconstruction algorithms often rely on batching events~\cite{Rebecq20pami,Stoffregen20eccv,Jiang20cvpr} and excessive unwanted flicker can dominate the data in a batch and lead to poor reconstructions of foreground activity.
Flicker is also a major issue in optical flow and other motion estimation methods that depend on image consistency.
Real-world flickering light sources are complex to model.
Rarely are they perfect sinusoidal signals due to the electronics that are used to generate the driving current and the frequency response usually has many strong harmonics \cite{frater2015light}.

In this paper, we propose an algorithm for event flicker removal.
To the best of the authors' knowledge, this is the first such algorithm that has been published.
A key contribution is the recognition that flicker events are highly non-sinusoidal and cannot be removed using a finite-dimensional notch filter.
This leads us to consider linear comb filters that attenuate all harmonics of a given frequency at the same time by feeding back a time-delayed version of the original signal.
We use both delayed feedback and delayed feed-forward~\cite{pei1998comb,jovanovic2005new} to design a linear filter response that is effective at passing the DC component of a signal while strongly attenuating all harmonics of a desired base frequency.
Since the design is based on a linear filter, there is a short transient while the internal filter state converges to attenuate the undesired signal.
The resulting algorithm is highly effective in removing flicker from event streams.
The algorithm is suitable for integration into a Field-Programmable Gate Array (FPGA) or Application-Specific Integrated Circuit (ASIC), allowing the algorithm to be built into the camera hardware.
In the present paper we provide only a demonstration implementation in C++ to demonstrate its potential.
In summary,
\begin{itemize}
	\item We propose a modified linear comb filter for event flicker removal.
	The algorithm allows parallel pixel-wise computation and is suitable for implementation on low-latency and low-power hardware circuits, such as FPGA and ASIC.
	
	
	\item We demonstrate event flicker removal performance in high frame rate and fast motion video reconstruction in the presence of fluorescent lighting.
\end{itemize}


\section{Related Works}
Since event cameras are prone to noise \cite{lichtsteiner2008128} many event data de-noising algorithms have been used in image reconstruction and de-blurring algorithms.
For example, noise in event data can be modelled by estimating event data bias and contrast threshold to match event data to reference intensity images~\cite{Brandli14iscas,wang2019threshold,wang2020asynchronous}.
To avoid accumulating event bias over time, the Complementary Filter (CF) \cite{Scheerlinck18accv} exponentially `forgets' historical event data using a single filter gain parameter.
Pan \textit{et~al.}~\cite{Pan20pami} estimates event camera contrast threshold by minimising energy function of image sharpness.
Some neural networks~\cite{Rebecq20pami,Stoffregen20eccv,Lin20eccv,Jiang20cvpr} rely on regularisation and spatial priors to reduce artefacts caused by event data noise.


In addition to estimating event sensitivity and priors, event filtering methods can remove noisy events as a preprocessing step for other algorithms.
Delbruck~\cite{Delbruck08issle} firstly introduced an effective background activity filter that directly removes isolated events with a low spatial-temporal correlation between the nearest neighbouring event data.
The algorithm has been implemented in the event-based sensor software jAER~\cite{jAER-software}.
Ieng \textit{et~al.}~\cite{ieng2014asynchronous} reduced noise by spatial-temporal filtering event data asynchronously and Liu \textit{et~al.}~\cite{liu2015design} implemented a background activity filter in hardware that filters out 98\% background noise with 10\% loss of real data.
Later, Padala \textit{et~al.}~\cite{padala2018noise} achieved higher signal to noise ratio
and Khodamoradi \textit{et~al.}~\cite{khodamoradi2018n} reduced memory complexity of the background activity filter from $\mathcal{O}(N^2)$ to $\mathcal{O}(N)$.
Apart from the correlation with the nearest neighbour, event data noise was also modelled based on differing polarity, refractory period and inter spike interval~\cite{Czech16biorob,wang2020asynchronous}.
Recently, {Wang \textit{et~al.}~\cite{wang2020joint} bridged event and image data via a noise-robust motion compensation model and proposed a joint filtering method.
	
To the authors' knowledge, there is no published algorithm that is targeted at attenuation of flicker events in event streams.
In contrast, flicker removal is a well-studied field for conventional cameras.
For conventional images, flicker is modelled as a dramatic intensity change between two neighbouring frames \cite{shiau2017vlsideisgn} rather than at a pixel level.
Chono \textit{et~al.}~\cite{chono2006detented} proposed a quantisation-based method to insert a periodically intra-coded pictures derived from the previous predictive-coded pictures, that suppresses the effect of flickering.
Guthier \textit{et~al.}~\cite{guthier2011flicker} proposed an iterative flicker removal algorithm that uses an average intensity change threshold to detect flicker and an iterative intensity scaling to compensate until variation falls below the threshold.
Shiau \textit{et~al.}~\cite{shiau2017vlsideisgn} designed an adjuster to the video frame, which estimates the atmospheric light of each frame, and assigned a multiplier based on that weight.
Many modern video cameras come with 50 or 60Hz de-flicker as a standard functionality.


\section{METHOD}
In this section we present the mathematical model of our event-based flicker removal algorithm.

\subsection{Mathematical Model}
\subsubsection{Event Camera Model}
As an asynchronous image sensor, the output of an event camera follows the Address Event Representation (AER) \cite{chan2007AER}.
The intensity change of the $i^{th}$ event $e_i$ is encoded in a vector $[t_i, p_i^x, p_i^y, \sigma_i]^T$ with event timestamp $t_i$, pixel coordinate $p_i = (p_i^x,p_i^y)$, and polarity $\sigma_i$ for positive or negative direction of the log-intensity change.
We model a single event as an impulse signal $\delta$ in continuous time
\begin{equation}
e_i(t)=\sigma_i c^{p_i} \delta(t - t_i), \hspace{0.1cm} i \in 1,2,3,\dots
\label{Eq:unit impulse}
\end{equation}
where $c^{p_i}$ is the contrast threshold at pixel $p_i$.
An event stream $E$ for asynchronous filter implementations is defined as
\begin{equation}
E(p,t) = \sum_{i=1}^{\infty}e_i(t)\delta_{p}(p_i) = \sum_{i=1}^{\infty}\sigma_i^p c^p \delta(t-t_i^p) \delta_{p}(p_i).
\label{Eq:impulse train}
\end{equation}
where $p$ is the pixel location considered and $t$ is the time index and $\delta_{p}(p_i)$ is the Kronecker delta function that is unity when $p = p_i$ and zero otherwise~\cite{hassani2008mathematical}.

\subsubsection{Flickering and Harmonics}
The high temporal resolution property enables event cameras to capture events that encode high-frequency flickering intensity as a periodic waveform.
Reproducing the complex waveform of a light source requires a large number of events per second even though the light source appears to be of a constant brightness for a typical frame-based camera and human eye.
Fluorescent and LED lights under Alternating Current (AC) power source can be modelled as a periodic signal over time $t$.
Expanding the signal in Fourier series yields
\begin{equation}
f(t) = \sum_{k=0}^{\infty}a_k \cos(k \omega_0 t ) + b_k \sin(k \omega_0 t),
\label{Eq:Fourier representation}
\end{equation}
for Fourier coefficients $a_k$ and $b_k$.
Due to the electronics in fluorescent and LED lights the higher order coefficients are generally quite large and the signal cannot be attenuated by a simple notch filter at the base frequency.
The power spectral density of the signal \eqref{Eq:Fourier representation} consists of an infinite sum of harmonics at multiples $k \omega_0$ of the base frequency.

\subsection{Proposed Comb Filters}

\subsubsection{Feed-forward Comb Filter}
The basic feed-forward comb filter \cite{smith2010physical} subtracts a time-delayed version of the signal to itself (see Fig.~\ref{Fig:feed-forward comb}).
If the time delay corresponds exactly to the period $\tau = 2 \pi/\omega_0$ of the flicker, then the resulting summed signal attenuates the flicker.
The corresponding system block diagram is shown in Fig.~\ref{Fig:feed-forward_comb_filter}(a).
The bode plot shown in Fig.~\ref{Fig:feed-forward_comb_filter}(b) displays periodic notches to the input signal corresponding not only to the base frequency $f_0 = \omega_0 / 2\pi$, but also to the harmonics of the flicker frequency.
The infinite harmonic notch filter is characteristic of the time delay action; the cancellation of periodic signals works equally well at multiple cycles of the base period $\tau = \frac{2 \pi}{k \omega_0}$ as it does for $k = 1$.
This infinite-dimensional nature of the comb filter is ideal for the suppression of harmonic noise characteristic of flicker in event data.

The Laplace transform of the feed-forward comb filter is given by \cite{smith2010physical}
\begin{equation}
Y(s) = (1-e^{-s\tau})X(s)
\label{Eq:feed-forward_comb}
\end{equation}
The magnitude of the impulse response is computed by substituting complex frequency variable $s$ with $j\omega$
\begin{align}
|H(j\omega)| &= |(1- \cos(\omega \tau) + \sin (\omega \tau)j| \notag\\
&= \sqrt{2-2 \cos(\omega\tau)}.
\label{Eq:impulse response}
\end{align}
However, as seen from the bode plot in Fig.~\ref{Fig:feed-forward_comb_filter}(b), there is a significant magnitude and phase distortion around the notches.
The basic feed-forward comb filter also attenuates the DC component of the signal.
In the proposed application this will annihilate static scene information and would compromise the performance of the filter.

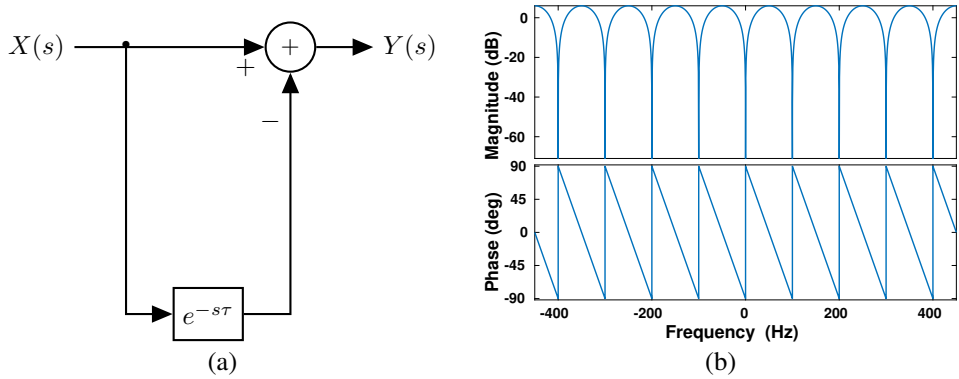
\begin{figure}[H]
	\centering
	\begin{tabular}{c c}
		\resizebox{0.34\linewidth}{!}{
			\begin{tikzpicture}[auto, thick, node distance = 1.6em, >=triangle 45]
			\draw
			node at (1.6, 0)(inputE){$X(s)$}
			node at (5, 0) [sum] (sum2) {\suma}
			node at (3.9, -3.6) [block] (delay) {$e^{-s\tau}$}
			node [right of = sum2, node distance = 1.6cm](out){$Y(s)$};
			
			\draw[->](inputE) -- node[pos = 0.9, yshift = -1.5em]{$+$}(sum2);
			\draw[->] (2.8, 0) |- (delay);
			\draw[->] (delay) -| node[pos = 0.9]{$-$}(sum2);
			\draw[->] (sum2) -- (out);
			
			\draw node at (2.8,0) {\textbullet};
			\end{tikzpicture}
		}
		&
		\includegraphics[width=0.35\linewidth]{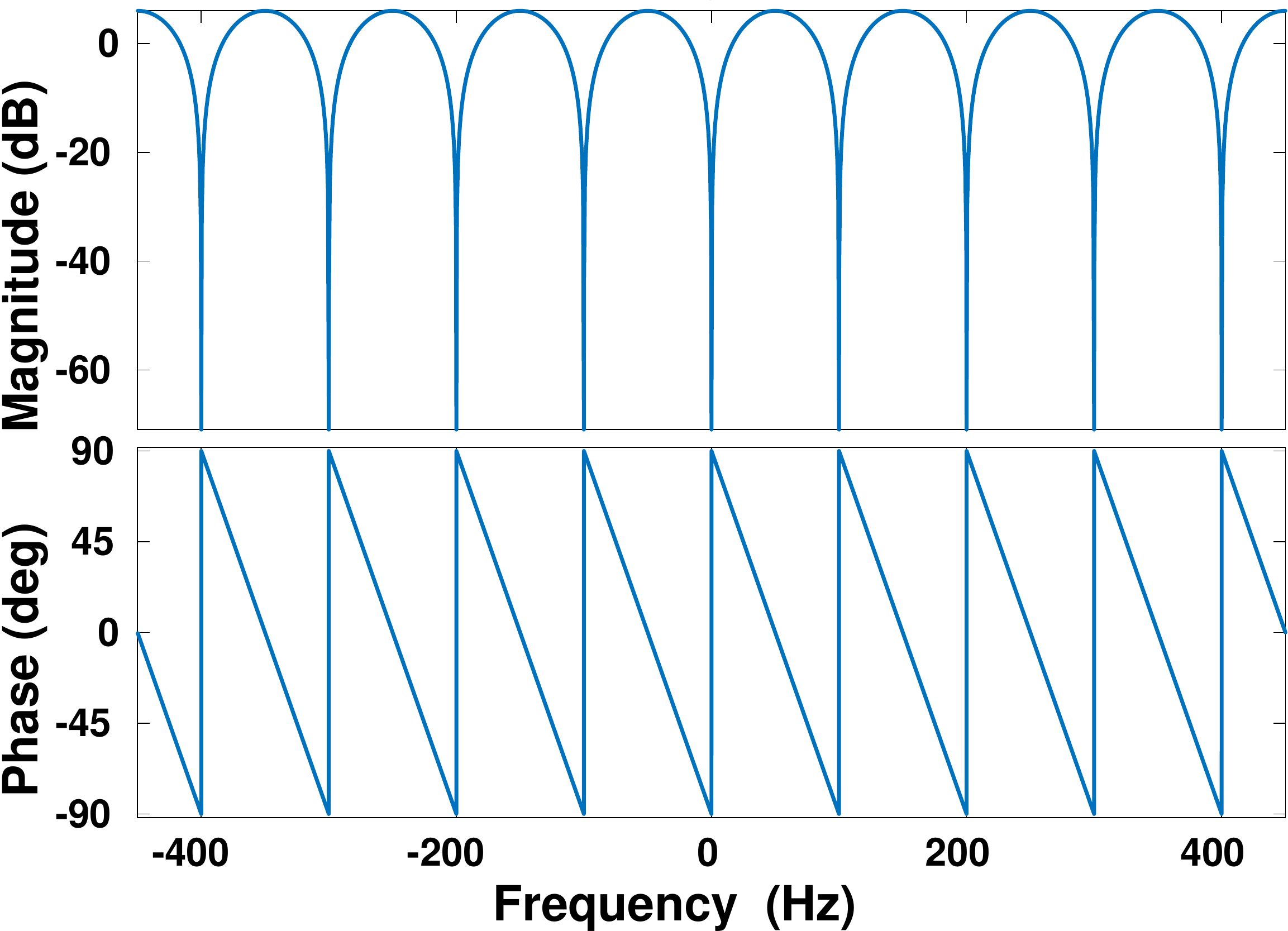} \\
		(a) & (b) \\
	\end{tabular}
	\caption{\label{Fig:feed-forward_comb_filter} 
		(a) Block diagram of the feed-forward comb filter (b) Bode plot of the feed-forward comb filter. The base frequency $f_0 = 1/\tau$ is set to 100Hz.
	}
	\label{Fig:feed-forward comb}
\end{figure}

\subsubsection{Distortion attenuation}
To solve the magnitude distortion problem of the feed-forward comb filter~\cite{pei1998comb}, we introduce a scaled positive feedback signal in addition to the feed-forward signal as shown in Fig.~\ref{Fig:pacf}.
The feed-forward comb filter can be thought of as a transfer function with an infinite set of zeros corresponding to the expansion of the exponential function.
The feedback loop introduces a corresponding infinite set of poles that are frequency aligned but have a slightly lower amplitude.
This introduces a near pole-zero cancellation of the filter response localising the frequency effect of the comb filter.
The transfer function for the feed-forward/feedback comb filter is
\begin{equation}
Y(s) = \frac{1-e^{-s\tau}}{1-\rho e^{-s\tau}}X(s),
\label{Eq:distortion1}
\end{equation}
where $0< \rho <1$ is a tuning parameter.
The parameter $\rho$ governs how closely the pole cancels the corresponding zero and determines the depth and width of each notch.
A higher $\rho$ value localises the filter effect close to the notch frequency and ensures that other frequencies are not distorted,
while a lower $\rho$ value widens both the notch and the associated frequency distortion up to the limit $\rho = 0$, where the basic feed-forward comb filter is recovered.
The corresponding bode plot is shown in Fig.~\ref{Fig:pacf}(b).

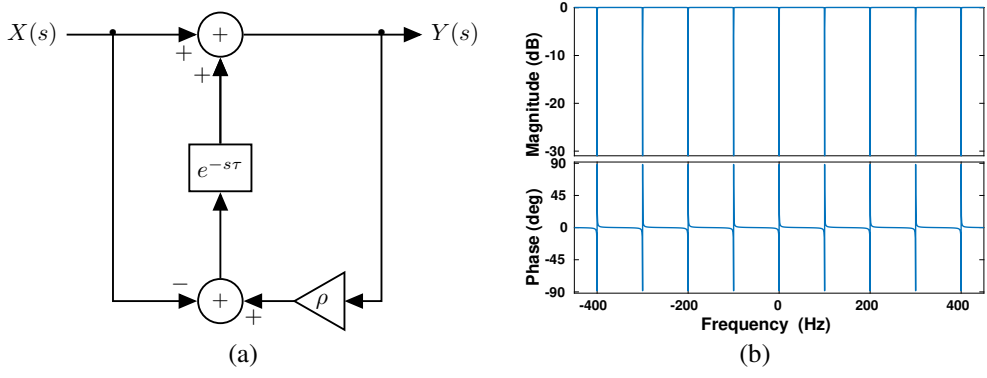
\begin{figure}[t!]
	\centering
	\begin{tabular}{c c}
		\resizebox{0.37\linewidth}{!}{
			\begin{tikzpicture}[auto, thick, node distance = 1.6em, >=triangle 45]
			\draw
			node at (2.3, 0)(inputE){$X(s)$}
			node at (5.1, 0)[sum](sum1){\suma}
			node at (8.6, 0)(out){$Y(s)$}
			node [block, below of = sum1](delay1){$e^{-s\tau}$}
			node [sum, below of = delay1](sum2){\suma}
			node [invgain, right of = sum2](gain1){$\rho$};
			
			\draw[->](inputE) -- node[pos = 0.9, yshift = -1.5em]{$+$}(sum1);
			\draw[->](sum1) -- (out);
			\draw[->](delay1) -- node[pos = 0.8]{$+$}(sum1);
			\draw[->](sum2) -- (delay1);
			\draw[->](delay1) -- (sum1);
			\draw[->](gain1) -- node[pos = 0.8]{$+$}(sum2);
			\draw[->](7.5,0) |- (gain1);
			\draw[->](3.5,0) |- node[pos = 0.9]{$-$}(sum2);
			\draw node at (7.5, 0) {\textbullet};
			\draw node at (3.5, 0) {\textbullet};
			\end{tikzpicture}
		}
		&
		\includegraphics[width=0.34\linewidth]{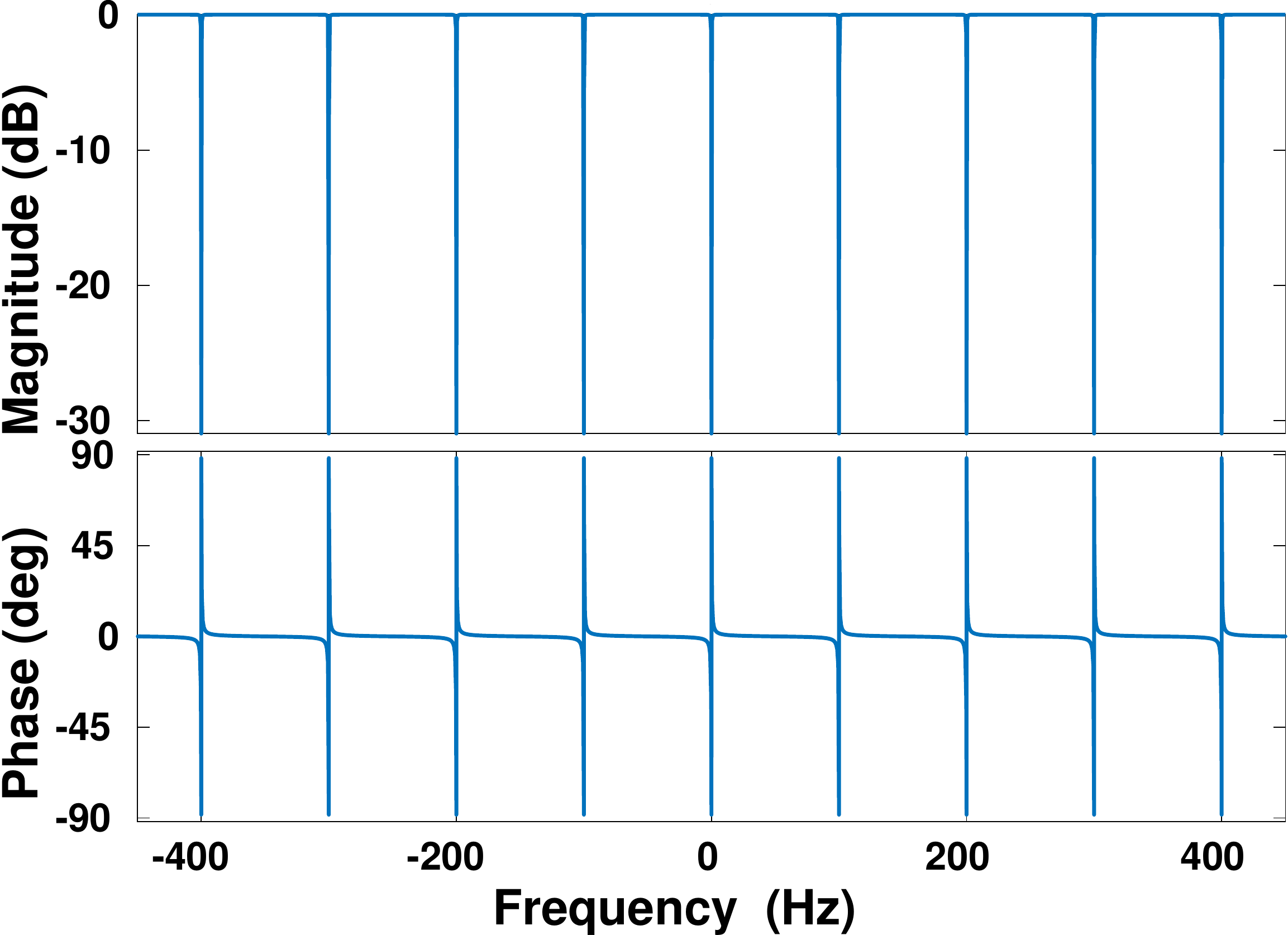} \\
		(a) &
		(b) \\
	\end{tabular}
	\caption{
		(a) Block diagram of the feed-forward/feedback comb filter (b) Bode plot of the feed-forward/feedback comb filter. $\rho$ can be set in range $0< \rho < 1$. We set $\rho = 0.6$ in this paper.
	}
	\label{Fig:pacf}
\end{figure}

\begin{figure*}[t!]
	\centering
	\begin{tabular}{c c}
		\resizebox{0.7\linewidth}{!}{
			\begin{tikzpicture}[auto, thick, node distance = 1.6cm, >=triangle 45]
			\draw
			node at (0,0) [name = int1]{$X(s)$}
			node [sum, right of = int1, node distance = 3cm](sum1){\suma}
			node [sum, right of = sum1, node distance = 6.4cm](sum2){\suma}
			node [block, right of= sum2, node distance = 3cm] (out){Sampler}
			node [right of = out, node distance = 2cm](eout){$E'(s)$}	
			node [block, below of = sum1] (delay1) {$e^{-s\tau_1}$}
			node [block, below of = sum2] (delay2) {$e^{-s\tau_2}$}
			node [sum, below of = delay1] (sum3){\suma}
			node [sum, below of = delay2] (sum4){\suma}
			node [invgain, right of = sum3](gain1){$\rho_1$}
			node [gain, left of = sum4](gain2){$\rho_2$};
			\draw[->](int1) -- node[pos = 0.9, yshift = 0em]{$+$}(sum1);
			\draw[->](sum1) -- node[pos = 0.9, yshift = 0em]{$+$}(sum2);
			\draw[->](sum2)--node[pos = 0.7, yshift = 0em]{$Y(s)$}(out);
			\draw[->](out)--(eout);
			\draw[->](delay1) -- node[pos = 0.9, yshift = 0em]{$+$}(sum1);
			\draw[->](delay2) -- node[pos = 0.9, yshift = 0em]{$+$}(sum2);
			\draw[->](sum3)--(delay1);
			\draw[->](sum4)--(delay2);
			\draw[->](gain1) -- node[pos = 0.8, yshift = 0em]{$+$}(sum3);
			\draw[->](gain2) -- node[pos = 0.8, yshift = 0em]{$-$}(sum4);
			\draw[->](1.4, 0) |- node[pos = 0.9, yshift = 0em]{$-$}(sum3);
			\draw[->](10.8, 0) |- node[pos = 0.9, yshift = 0em]{$+$}(sum4);
			\draw[->](6.2, 0) |- (gain1);
			\draw[->](6.2, 0) |- (gain2);
			\draw node at (1.4,0) {\textbullet};
			\draw node at (10.8, 0) {\textbullet};
			\draw node at (6.2, 0) {\textbullet};
			\end{tikzpicture}
		} \hspace{-80mm} & \\
		(a) Block diagram \hspace{-65mm} & \vspace{5mm} \\ 
		\hspace{-10mm}
		\resizebox{0.25\hsize}{!}{\begin{tikzpicture}[scale=2]
			\draw [-latex] (-1.5, 0) -- (1.5, 0) node [above left] {$\Re$};
			\draw [-latex] (0, -1.5) -- (0, 1.5) node [below right] {$\Im$};
			
			\draw[blue, dashed] (0,1) arc (90:-270:1);
			\draw node at (36:1) [solid, fill=white, circle,scale = 0.6, draw=black]{};
			\draw node at (72:1) [solid, fill=white, circle,scale = 0.6, draw=black]{};
			\draw node at (108:1) [solid, fill=white, circle,scale = 0.6, draw=black]{};
			\draw node at (144:1) [solid, fill=white, circle,scale = 0.6, draw=black]{};
			\draw node at (180:1) [solid, fill=white, circle,scale = 0.6, draw=black]{};
			\draw node at (216:1) [solid, fill=white, circle,scale = 0.6, draw=black]{};
			\draw node at (252:1) [solid, fill=white, circle,scale = 0.6, draw=black]{};
			\draw node at (288:1) [solid, fill=white, circle,scale = 0.6, draw=black]{};
			\draw node at (324:1) [solid, fill=white, circle,scale = 0.6, draw=black]{};
			\draw node at (36:0.85) [solid, cross out, scale = 0.6, draw=black]{};
			\draw node at (72:0.85) [solid, cross out, scale = 0.6, draw=black]{};
			\draw node at (108:0.85) [solid, cross out, scale = 0.6, draw=black]{};
			\draw node at (144:0.85) [solid, cross out, scale = 0.6, draw=black]{};
			\draw node at (180:0.85) [solid, cross out, scale = 0.6, draw=black]{};
			\draw node at (216:0.85) [solid, cross out, scale = 0.6, draw=black]{};
			\draw node at (252:0.85) [solid, cross out, scale = 0.6, draw=black]{};
			\draw node at (288:0.85) [solid, cross out, scale = 0.6, draw=black]{};
			\draw node at (324:0.85) [solid, cross out, scale = 0.6, draw=black]{};
			\end{tikzpicture}} & 
		\includegraphics[width=0.33\linewidth]{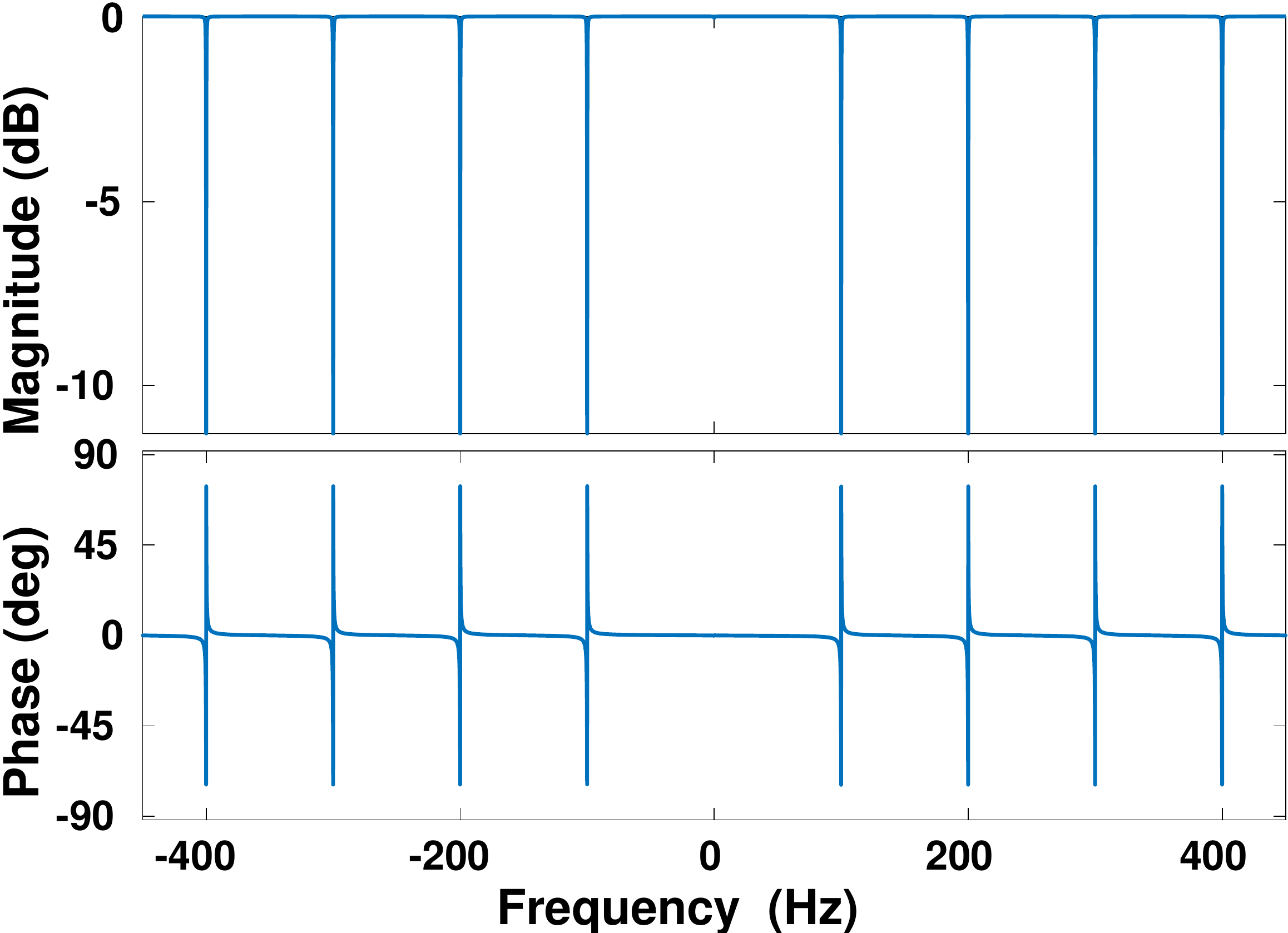} \\ 
		(b) Pole-zero plot & (c) Bode diagram   \\
	\end{tabular}
	\caption{\label{fig:pacf and acf}
		Our proposed event-based flicker removal algorithm.
	}
\end{figure*}
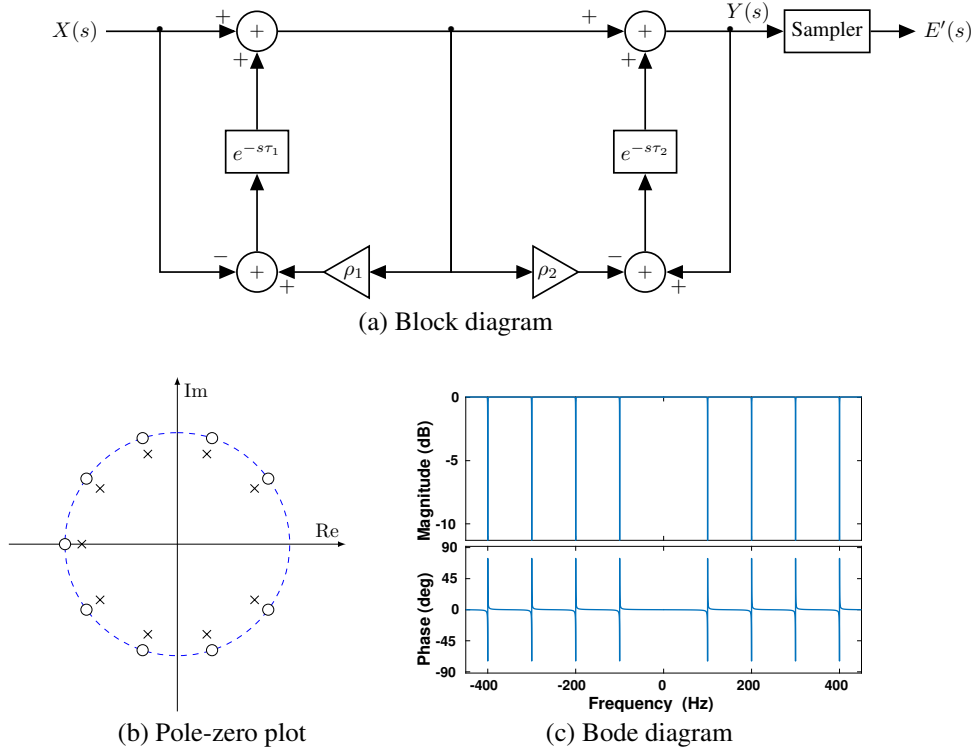

The combination of a feed-forward and feedback filter provides an infinite sequence of notch filters at all harmonics of a base frequency with a tunable factor $\rho$ that allows the notch to be widened at the cost of increased frequency distortion.
The remaining difficulty is the fact that the feed-forward loop still cancels the DC response of the base signal.

\subsubsection{Proposed event-based flicker removal algorithm}

The proposed filter (Figure~\ref{fig:pacf and acf}) uses a reciprocal comb filter~\cite{jovanovic2005new} (in resonant mode with gain on the feed-forward path allowing the feedback to dominant and emphasise flicker) at a much higher frequency with a smaller time-constant
$\tau_2 << \tau_1$ to introduce a cancellation of the DC pole-zero notch filter.
The Laplace transform of the filter is
\begin{equation}
Y(s) = \frac{1-e^{-s\tau_1}}{1-\rho_1\cdot e^{-s\tau_1}}\cdot\frac{1-\rho_2 e^{-s\tau_2}}{1-e^{-s\tau_2}}X(s),
\label{Eq:ACF}
\end{equation}
where the gain $\rho_2$ serves the same purpose as $\rho_1$ in \eqref{Eq:distortion1}. 
The time constant $\tau_2$ must be chosen significantly smaller than $\tau_1$, in our work we typically use $\tau_2 = \tau_1/10$ and tune the gain $\rho_2$ to satisfy $\tau_2 (1-\rho_1) = \tau_1(1-\rho_2)$ to ensure that the decade slope of the resonant mode matches the attenuation of the DC notch that needs to be cancelled.
The pole-zero plot in Fig.~\ref{fig:pacf and acf} provides a visualisation of the resulting pole zero cancellations.
Fig.~\ref{fig:pacf and acf}(c) shows the bode plot of our proposed filter with the desired notch filters at all harmonics with the DC component unaffected.

The time-domain solution of \eqref{Eq:ACF} can be written as
\begin{equation}
\resizebox{0.6\hsize}{!}{
	$\begin{split}
	y(t) &= x(t)- x(t-\tau_1) -\rho_2\cdot x(t-\tau_2)+\rho_2\cdot x(t-(\tau_1 + \tau_2))\\
	&+\rho_1\cdot y(t-\tau_1)+y(t-\tau_2)-\rho_1\cdot y(t-(\tau_1 + \tau_2)). \\
	\end{split}
	$
}
\label{Eq:ACF_time}
\end{equation}
This is a causal function that only depends on the present and past value of input $x$ and output $y$.
The resulting filter can be implemented by processing incoming events and keeping a ring buffer of events to implement the time-delays.

An event sampler is applied at the output to regenerate an event stream.
This sampler keeps a reference value of intensity and compares the output to the reference generating an event and resetting the reference when the filter output change is above the threshold.
The input signal to this sampler is a zero-order hold signal and the threshold need only be checked when an event occurs, either in the input signal or from one of the ring buffers that instantiates the time delay.
This can be implemented in digital logic.

\section{Experiment} \label{sec:experiment}

\subsection{Implementation details}
The filter is implemented continuously using C++.
It takes around 0.1 seconds to converge for a 50Hz base frequency (see Fig.~\ref{fig:filter converge}) corresponding to around 3-5 cycles of the flicker.
The comb filter parameters are set to: $\rho_1 = 0.6$, $\rho_2 = 1 - (1 - \rho_1) / 10$, $\tau_2 = 1/50$ for 50Hz base frequency, $\tau_2$ = $\tau_1/10$.

\subsection{Datasets}
To the best of the authors' knowledge, there is no existing open-source datasets and algorithm available targeted at the event camera flicker removal problem.
To fill the gap, we collected a set of real data under a flickering fluorescent light using a \textit{Prophesee Gen3} event camera (resolution 480 $\times$ 640).
Our datasets include fast motions of a basketball, table tennis, badminton and Nerf gun.
The flickering light source in the datasets is supplied by 50Hz alternating current in Australia, but the flickering primarily occurs at 100Hz because older fluorescent lights with magnetic ballasts flicker at twice the supply frequency~\cite{wilkins2010led} and 50Hz flickering mainly occurs at the two edges of the fluorescent light.

\begin{figure*}
	\newcommand{\colwidth}{3.8cm}
	\renewcommand{\tabcolsep}{0.4mm}
	\centering
	\resizebox{1\textwidth}{!}{
		\begin{tabular}
			{
				>{\centering\arraybackslash}m{4mm} >{\centering\arraybackslash}m{\colwidth}
				>{\centering\arraybackslash}m{\colwidth} >{\centering\arraybackslash}m{\colwidth}
				>{\centering\arraybackslash}m{\colwidth}
				>{\centering\arraybackslash}m{\colwidth}}
			\rotatebox{90}{\small (a) E2VID}
			&
			\includegraphics[width=\linewidth]{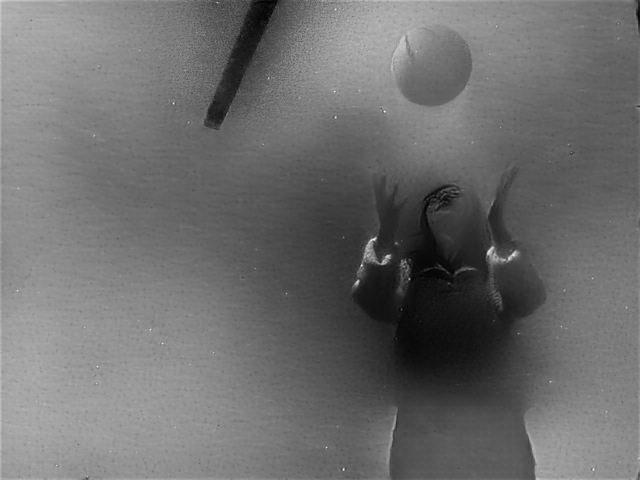}
			&
			\includegraphics[width=\linewidth]{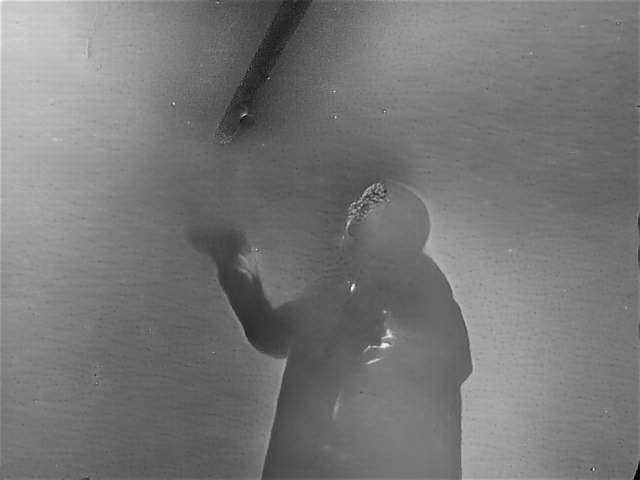}
			&
			\includegraphics[width=\linewidth]{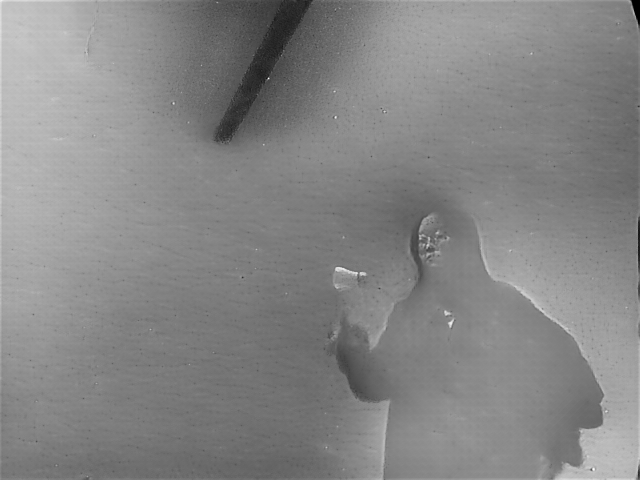}
			&
			\includegraphics[width=\linewidth]{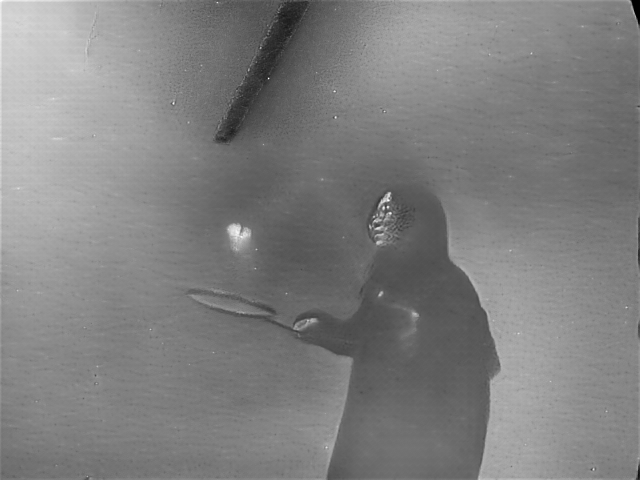}
			&
			\includegraphics[width=\linewidth]{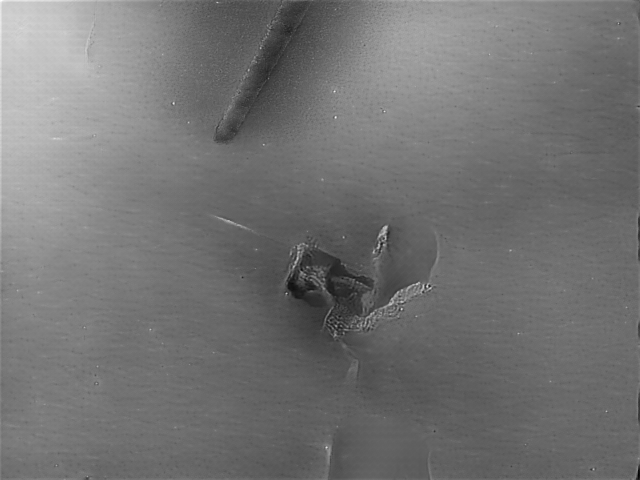}
			\\
			\rotatebox{90}{\small (b) EFR + E2VID}
			&
			\includegraphics[width=\linewidth]{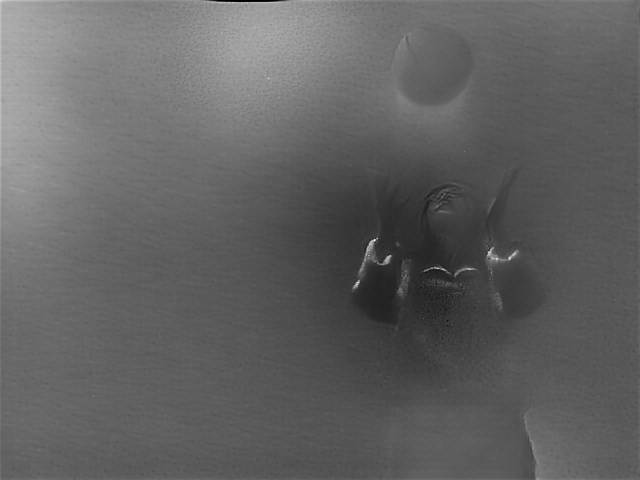}
			&
			\includegraphics[width=\linewidth]{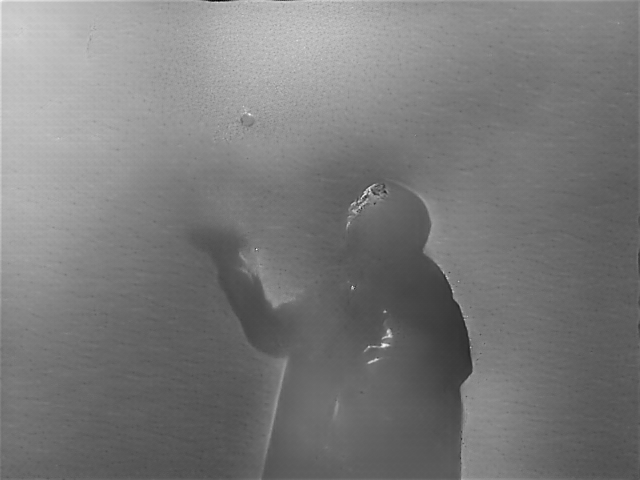}
			&
			\includegraphics[width=\linewidth]{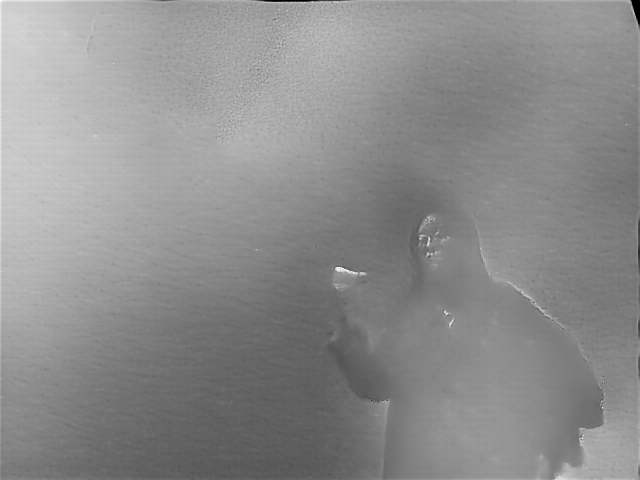}
			&
			\includegraphics[width=\linewidth]{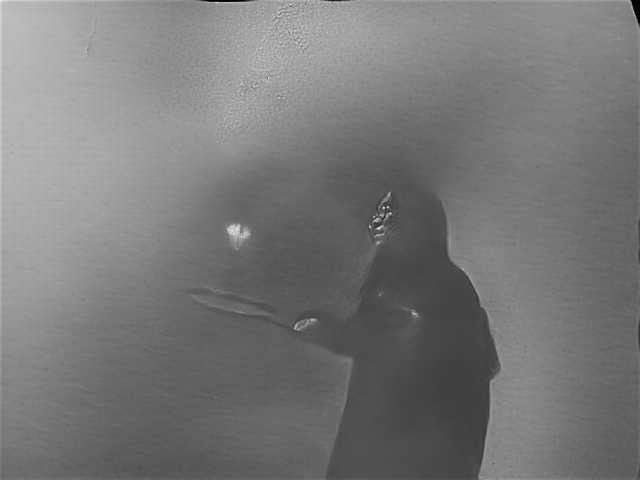}
			&
			\includegraphics[width=\linewidth]{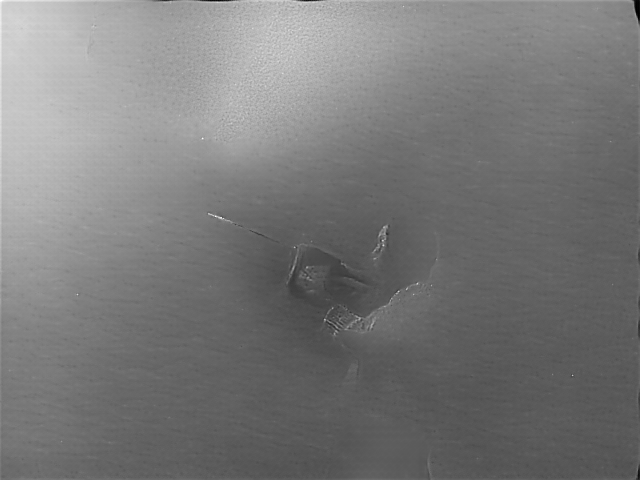}
			\\
			\rotatebox{90}{\small (c) Heatmap E2VID}
			&
			\includegraphics[width=\linewidth]{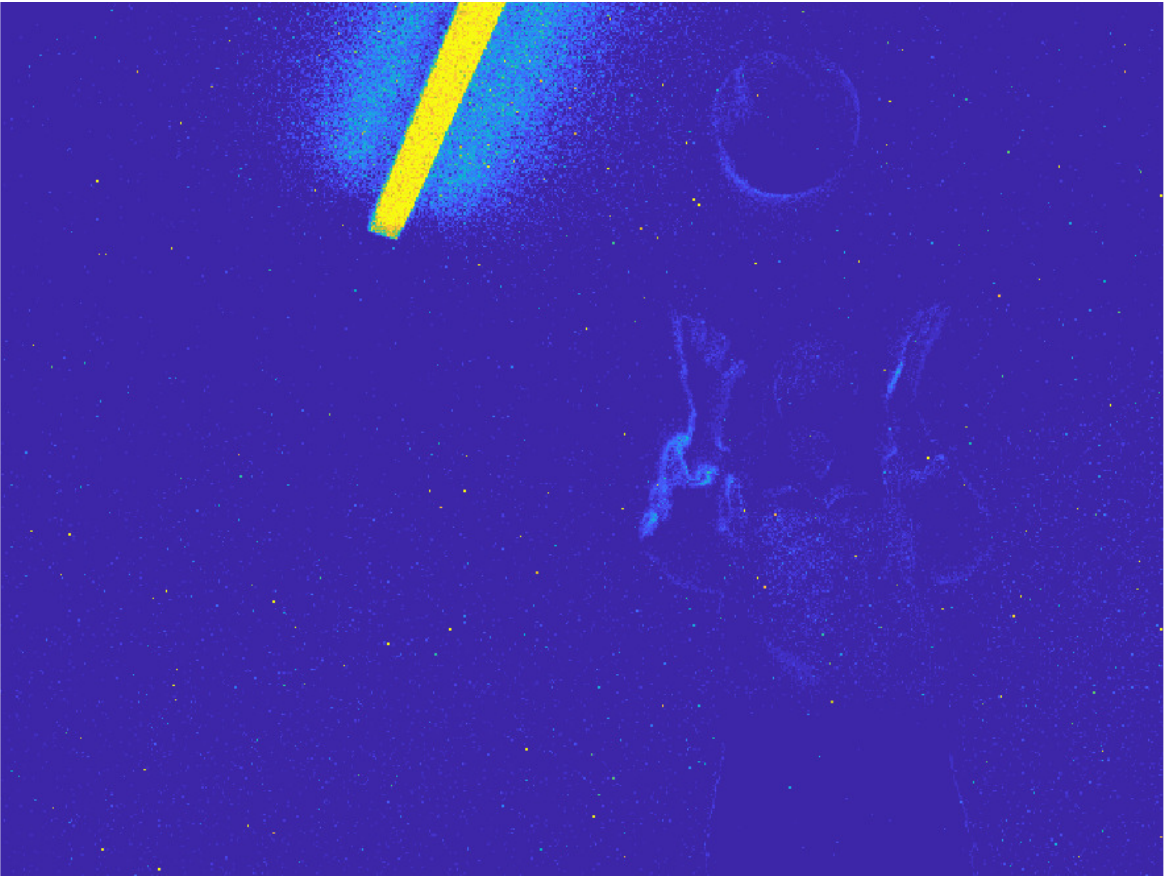}
			&
			\includegraphics[width=\linewidth]{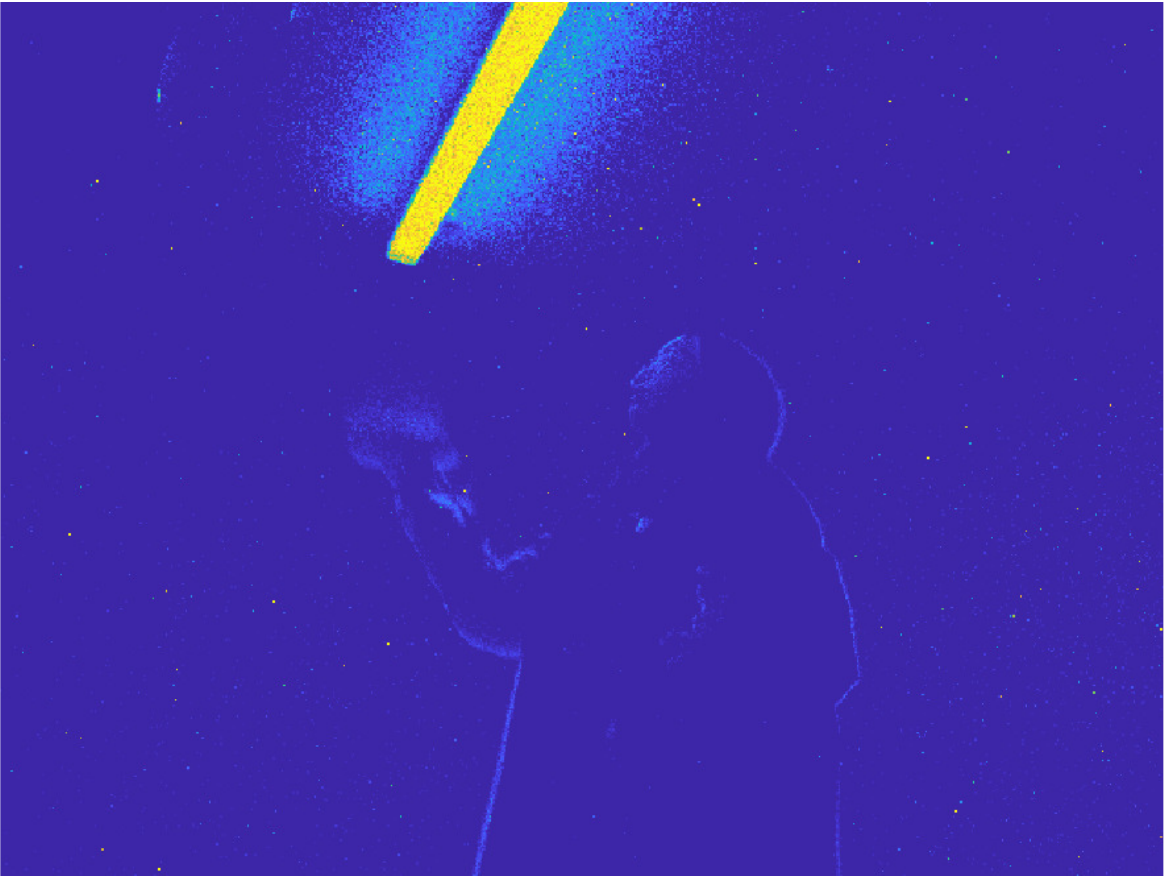}
			&
			\includegraphics[width=\linewidth]{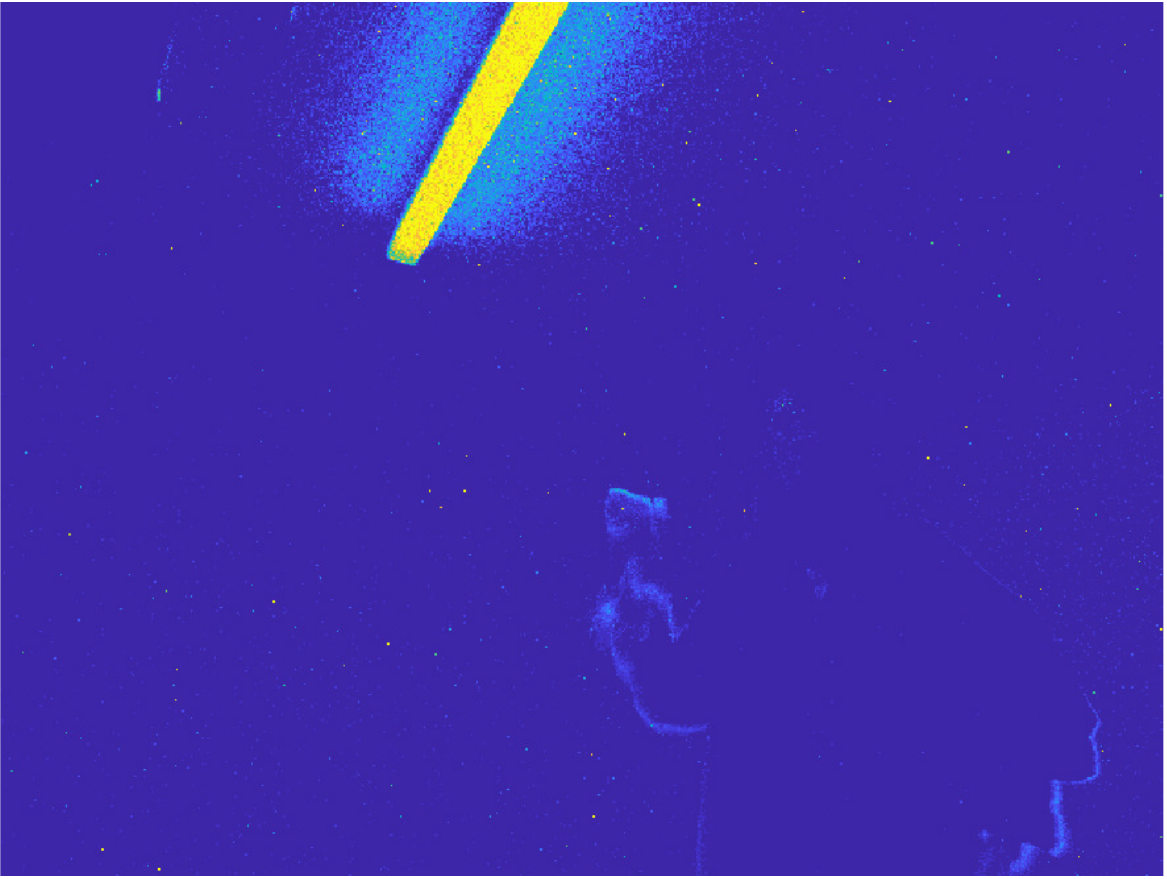}
			&
			\includegraphics[width=\linewidth]{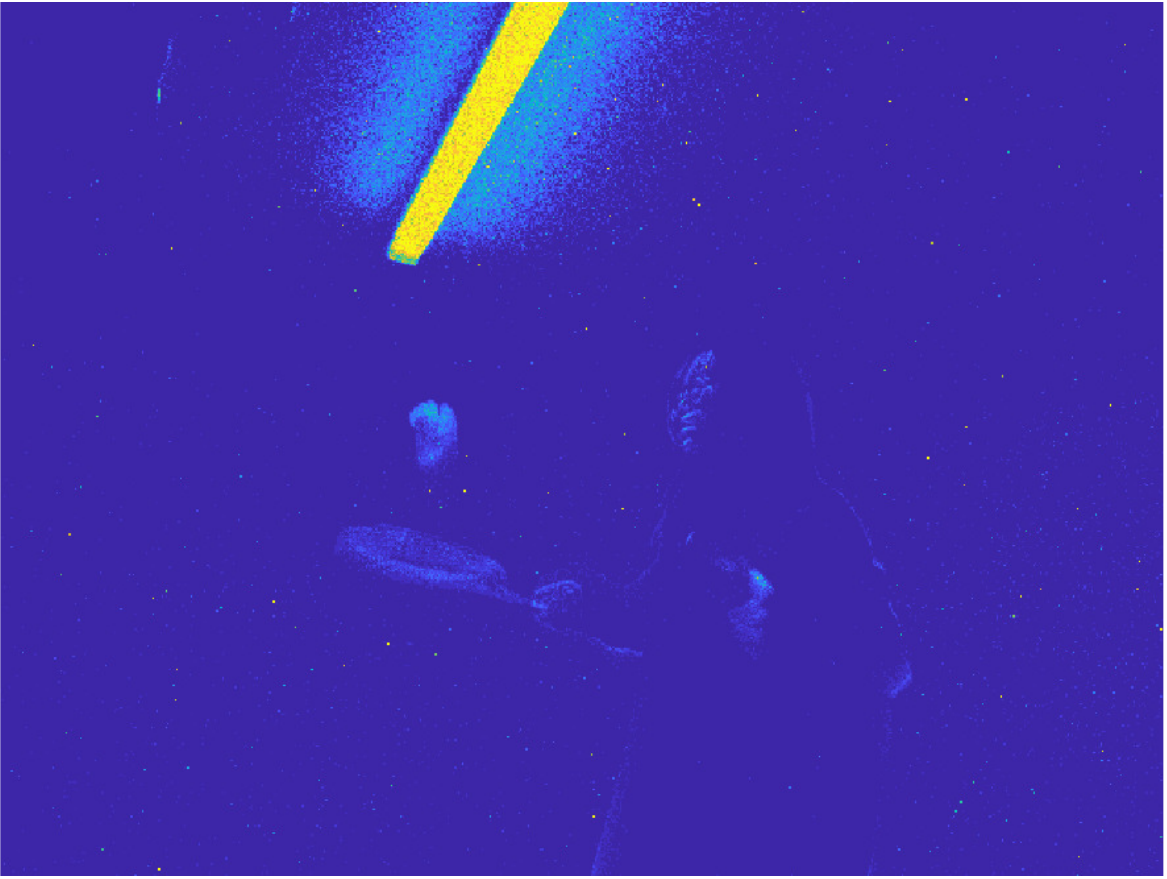}
			&
			\includegraphics[width=\linewidth]{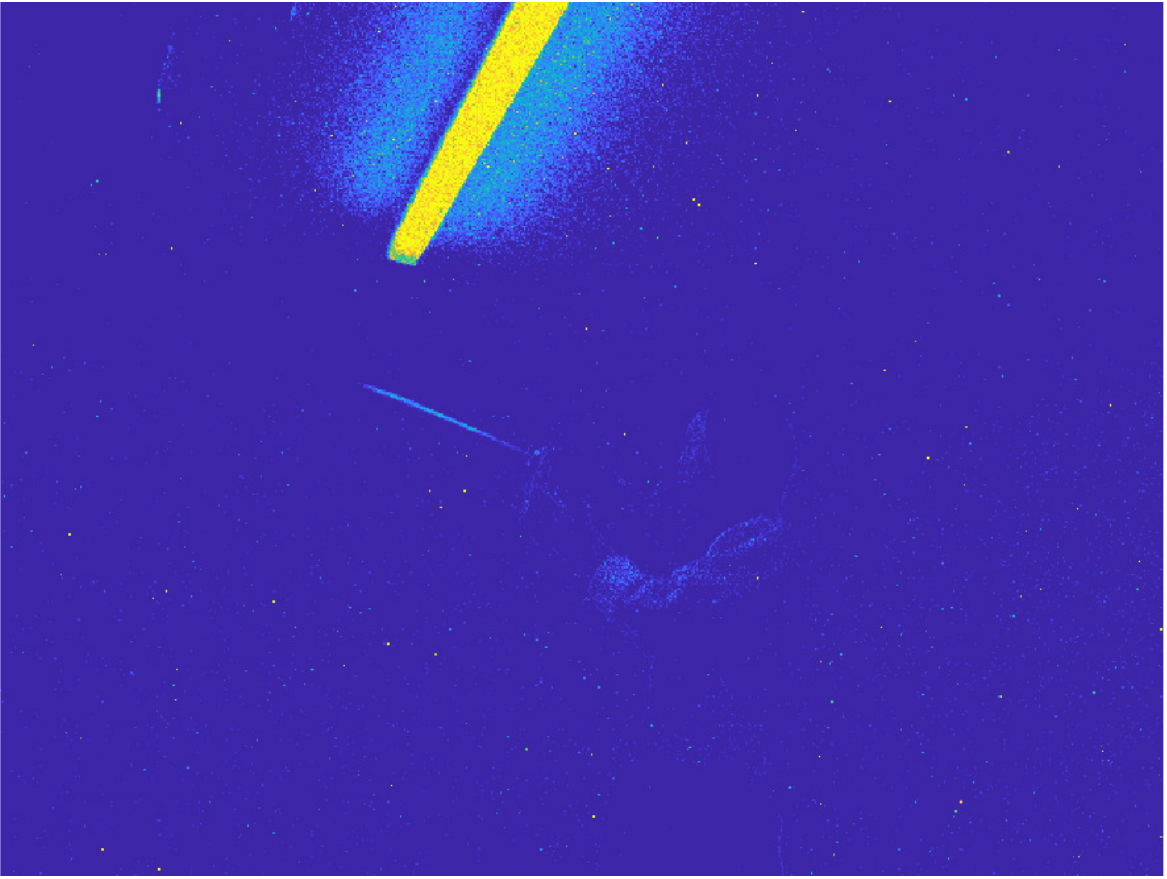}
			\\	
			\rotatebox{90}{\small (d) Heatmap EFR + E2VID}
			&
			\includegraphics[width=\linewidth]{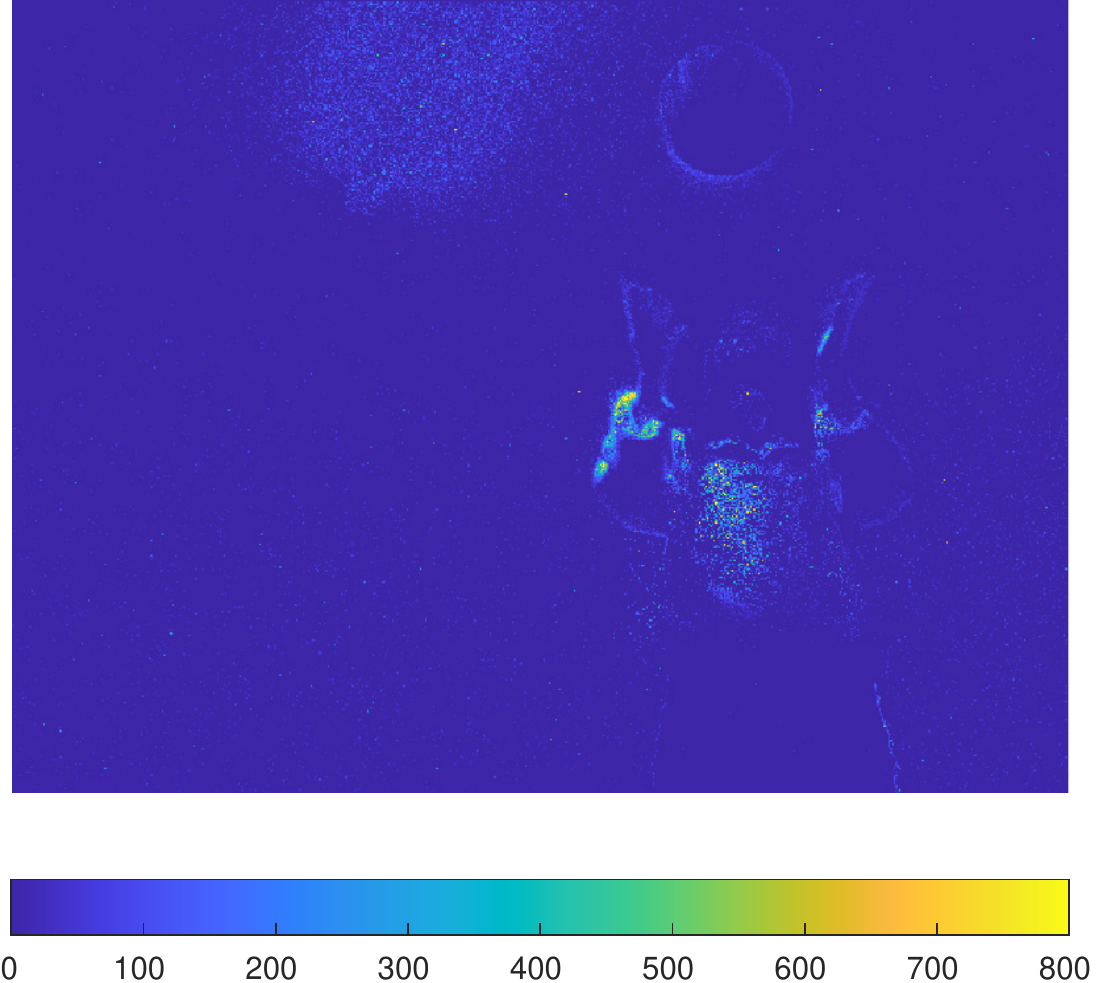}
			&
			\includegraphics[width=\linewidth]{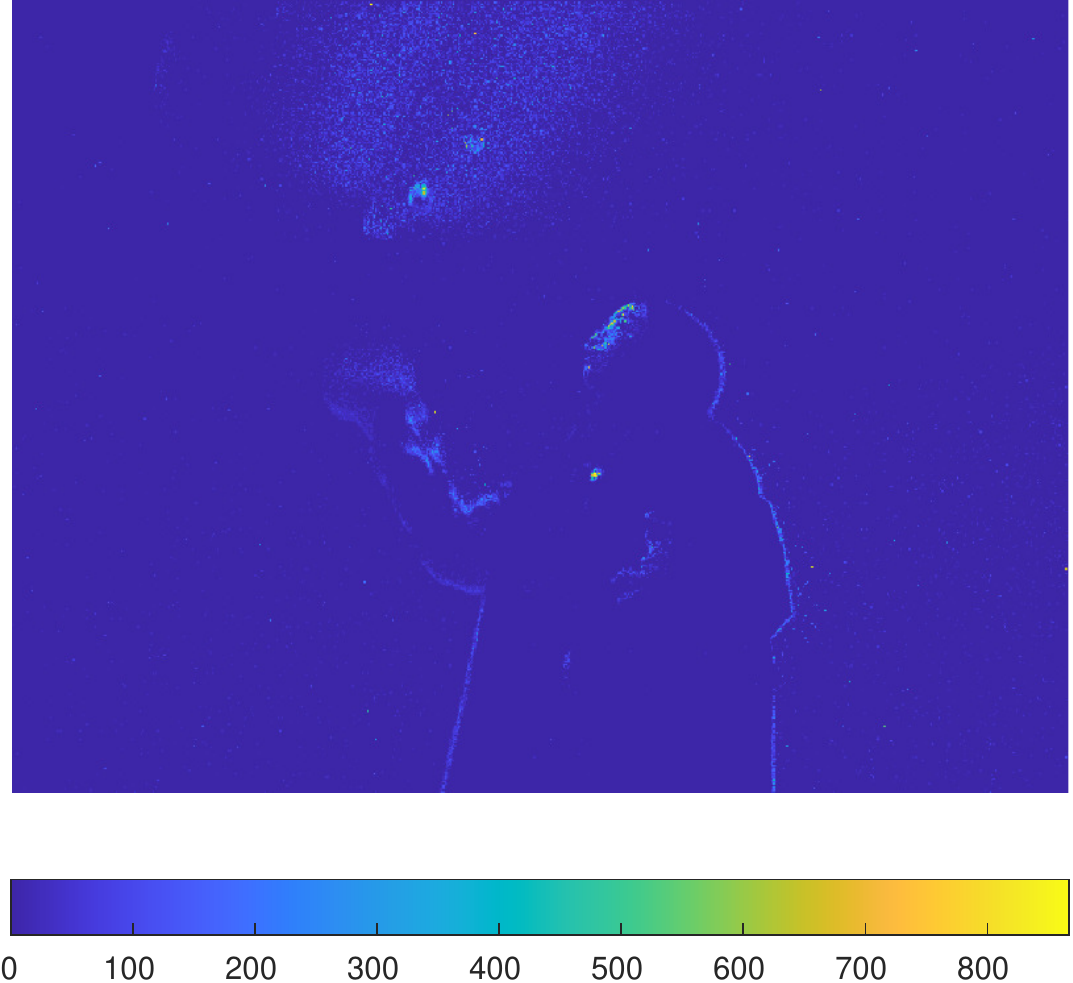}
			&
			\includegraphics[width=\linewidth]{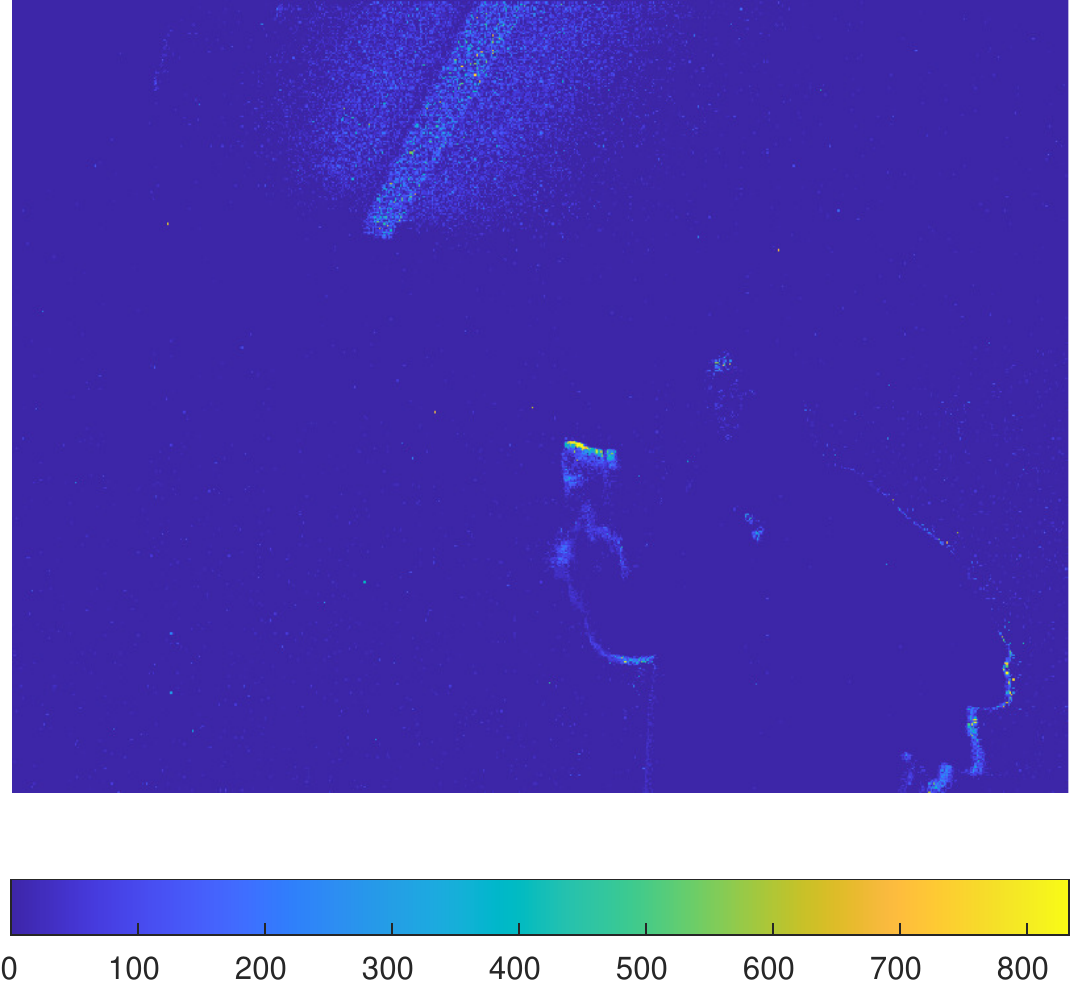}
			&
			\includegraphics[width=\linewidth]{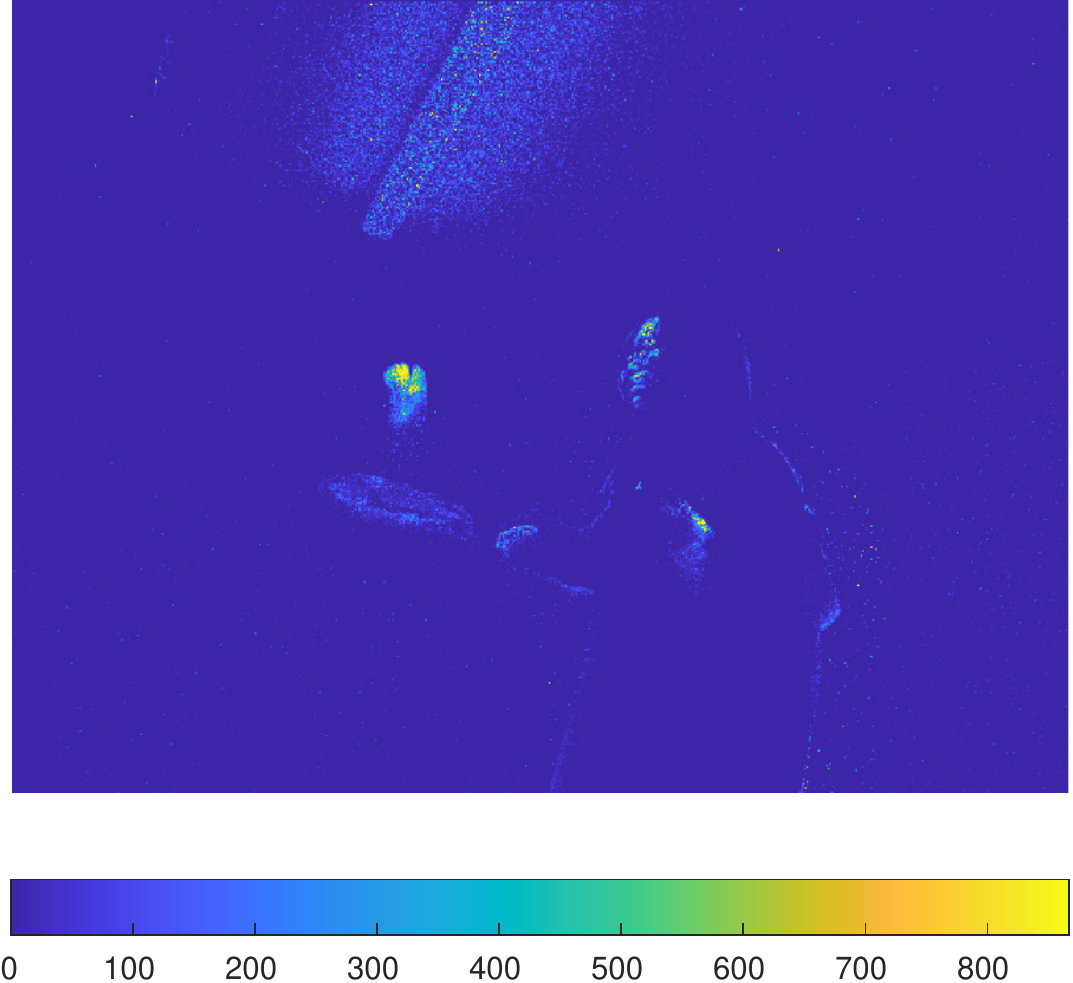}
			&
			\includegraphics[width=\linewidth]{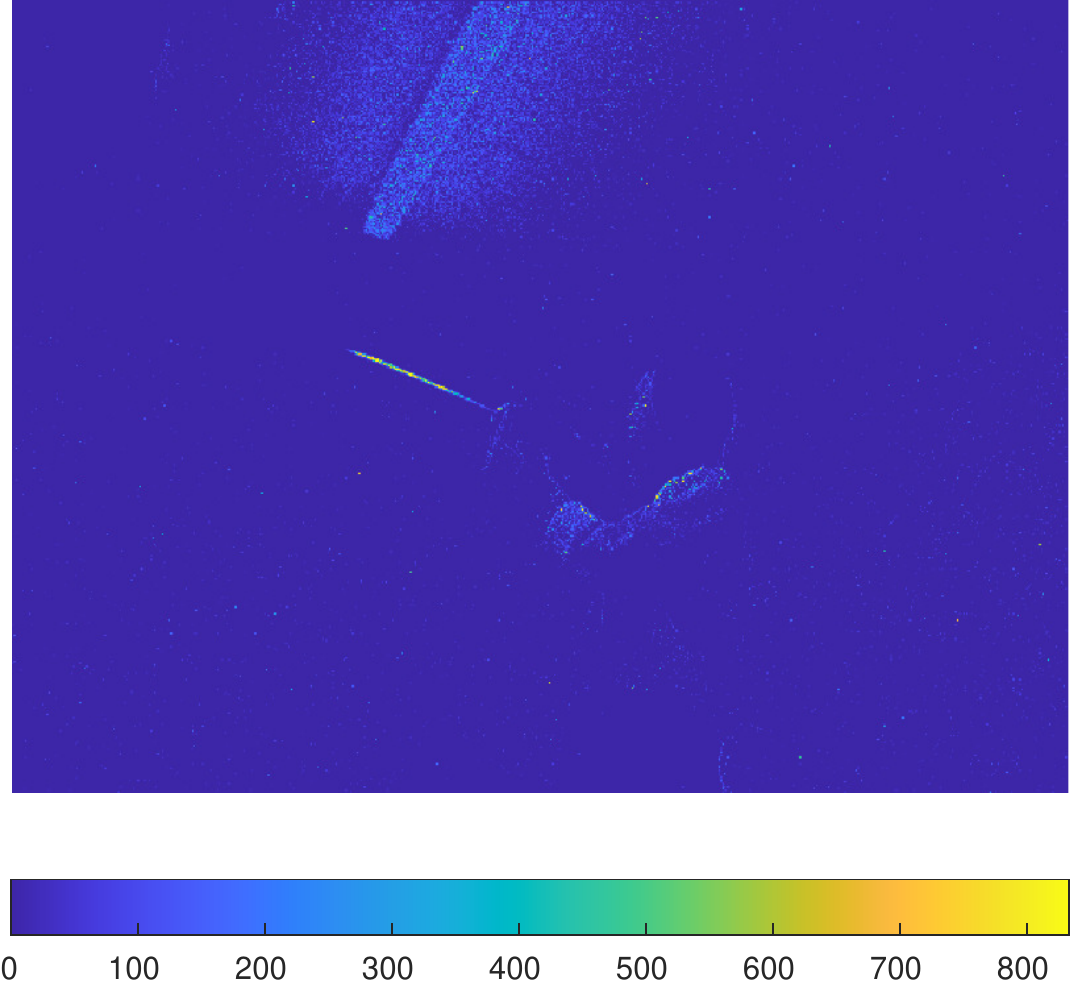}
			\\			
			& \texttt{Basketball}
			& \texttt{Table Tennis}
			& \texttt{Badminton 1}
			& \texttt{Badminton 2}
			& \texttt{Nerf Gun}
		\end{tabular}
	}
	\caption{
		Qualitatively analysis of the raw and filtered event data from five high-speed datasets with a flickering fluorescent light source.
		The first row (a) are image reconstruction E2VID~\cite{Rebecq20pami} using raw event data.
		The second raw (b) shows image reconstruction E2VID~\cite{Rebecq20pami} after event flicker removal using our proposed filter (EFR), and the flickering light source is clearly attenuated.
		The third and fourth row demonstrate heat maps of events per second of the raw and filtered event stream.
		The events rate from 0 to the maxima is coloured from blue to yellow.
		Before flicker removal, most events occur around the fluorescent light source.
		The ratio of events occurring at the flickering region to all events before filtering is 84.0\% on average.
		Our event-based flicker removal algorithm filters out most of the flickering events and mostly preserves the useful event data under the lights.
		Our filter is not affected by the motion across the flickering source (shown in \texttt{Table Tennis}).
	}\label{fig:e2vid}
	\vspace{-2mm}
\end{figure*}

\begin{figure}[t!]
	\centering
		\begin{tabular}{ c c }
			\\	\includegraphics[width=0.35\linewidth]{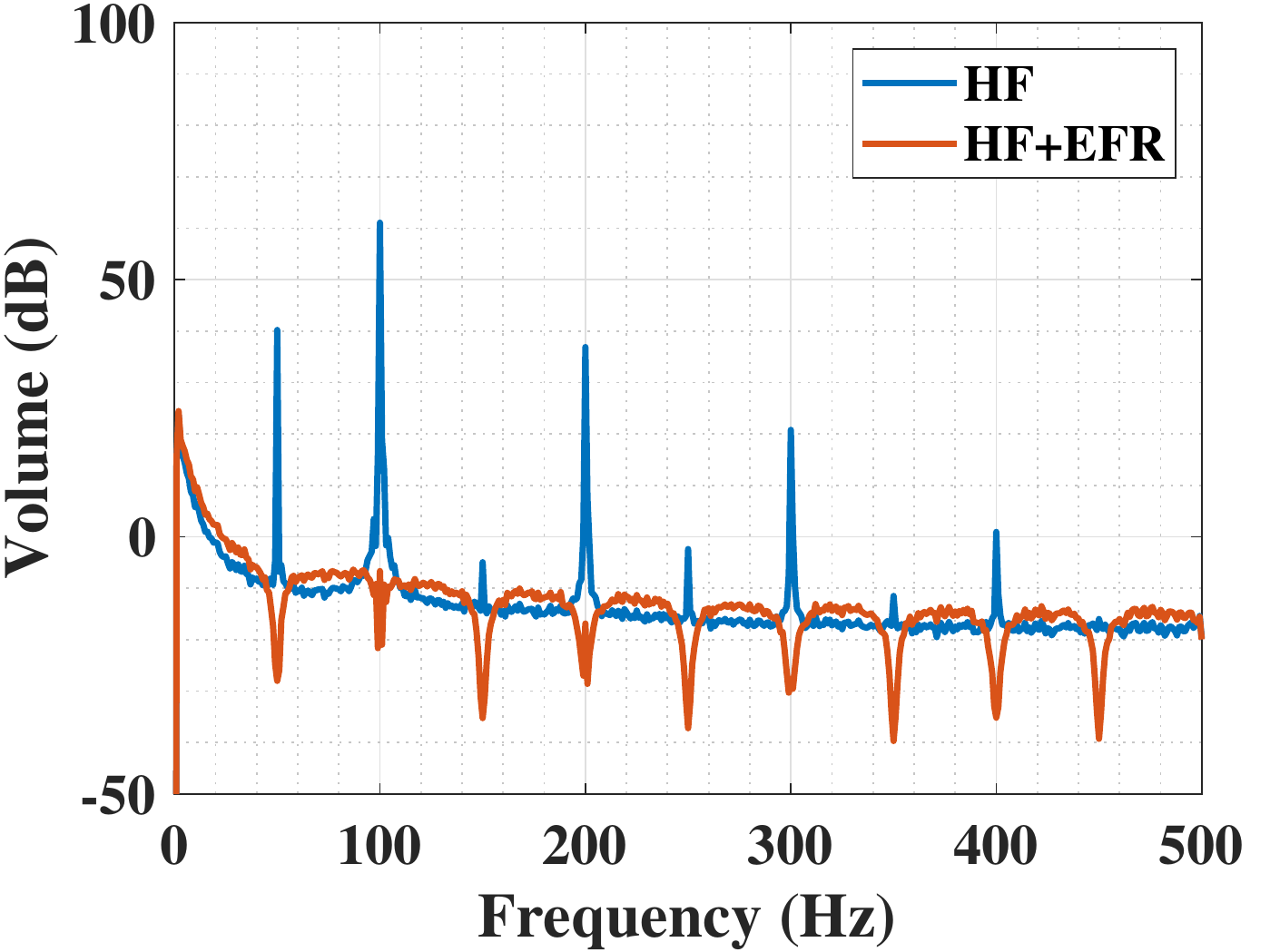} &
			\includegraphics[width=0.35\linewidth]{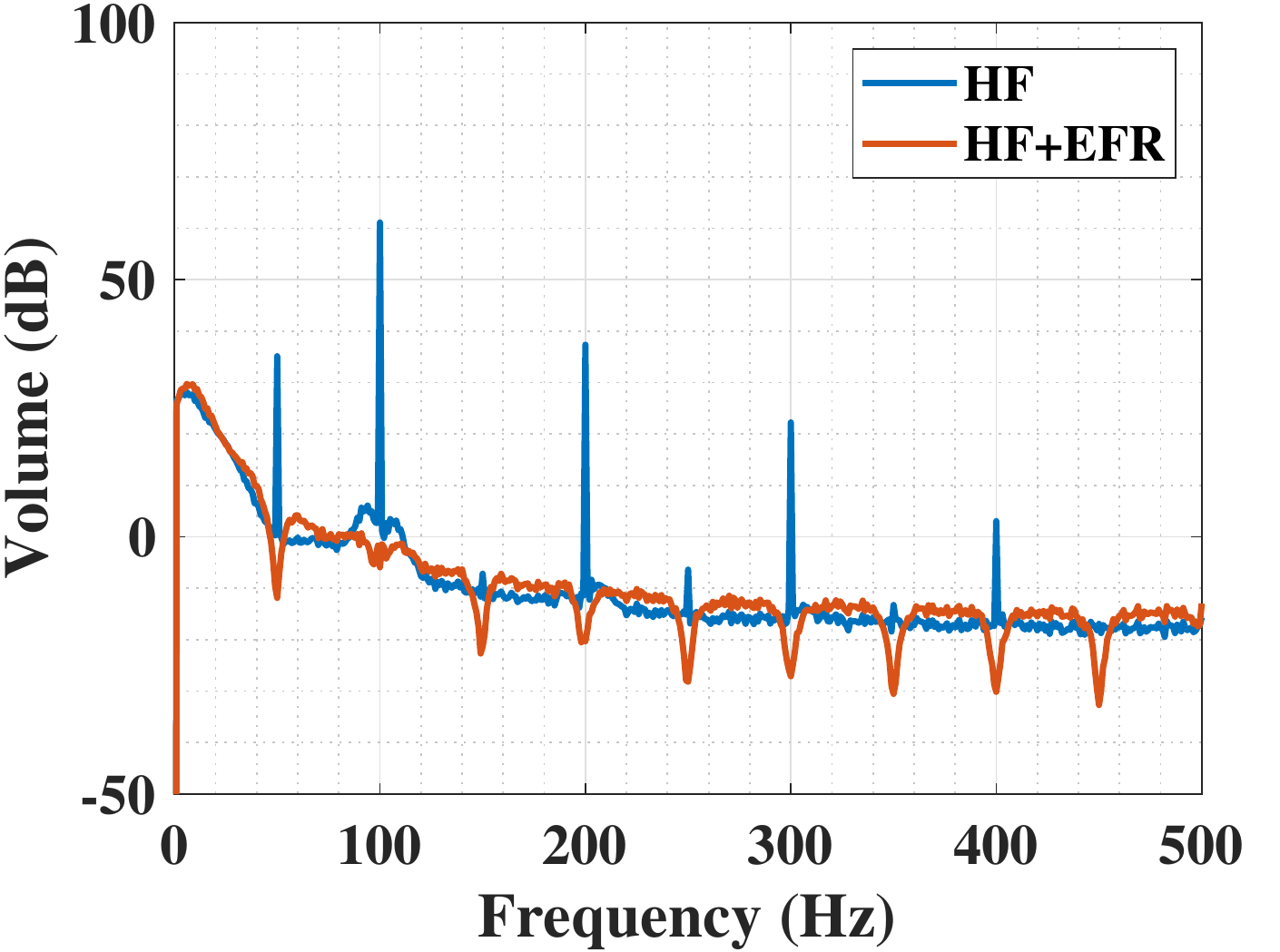} \\
			(a) \small \texttt{Basketball} & (b) \small \texttt{Table Tennis} \\
			\includegraphics[width=0.35\linewidth]{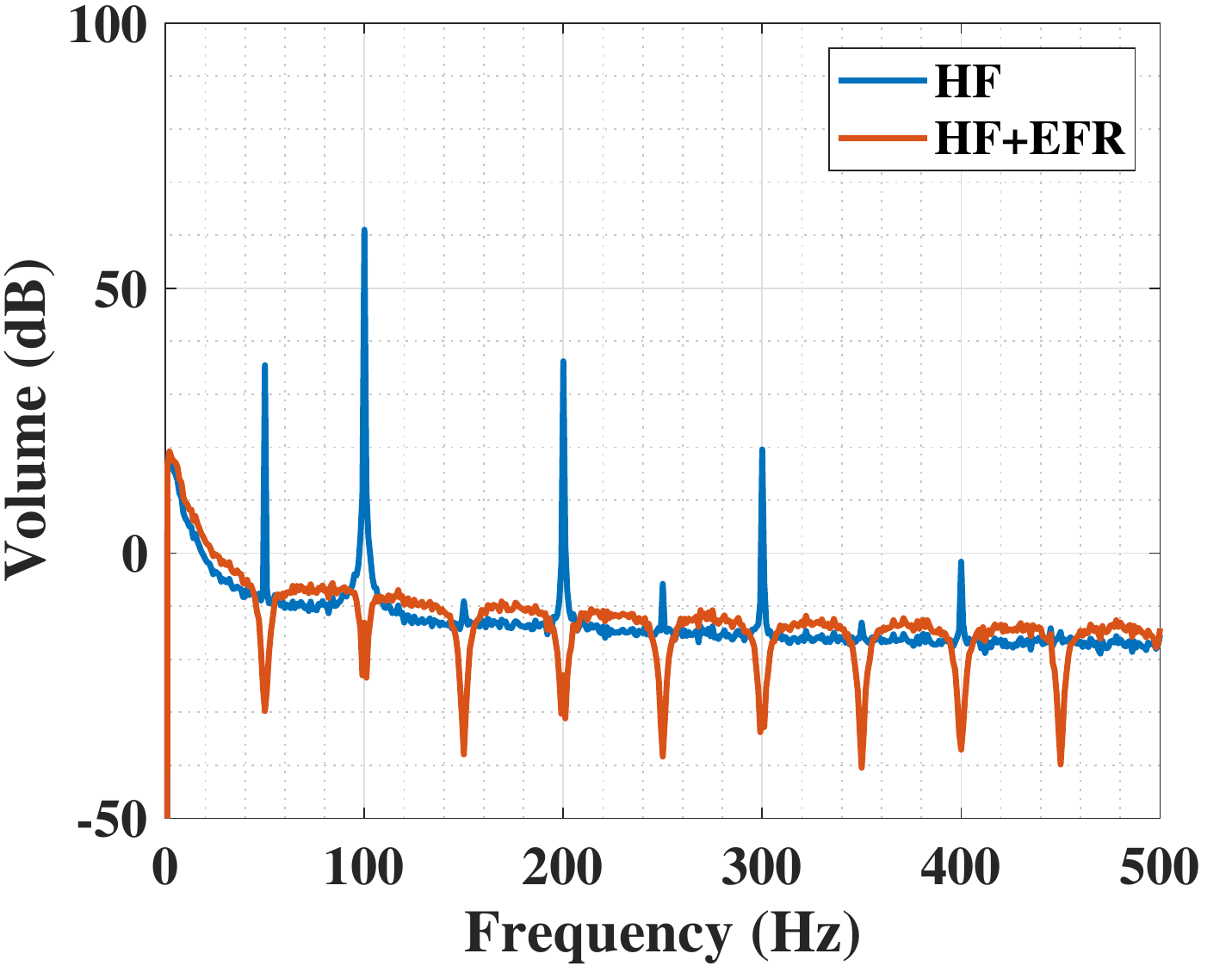} &
			\includegraphics[width=0.35\linewidth]{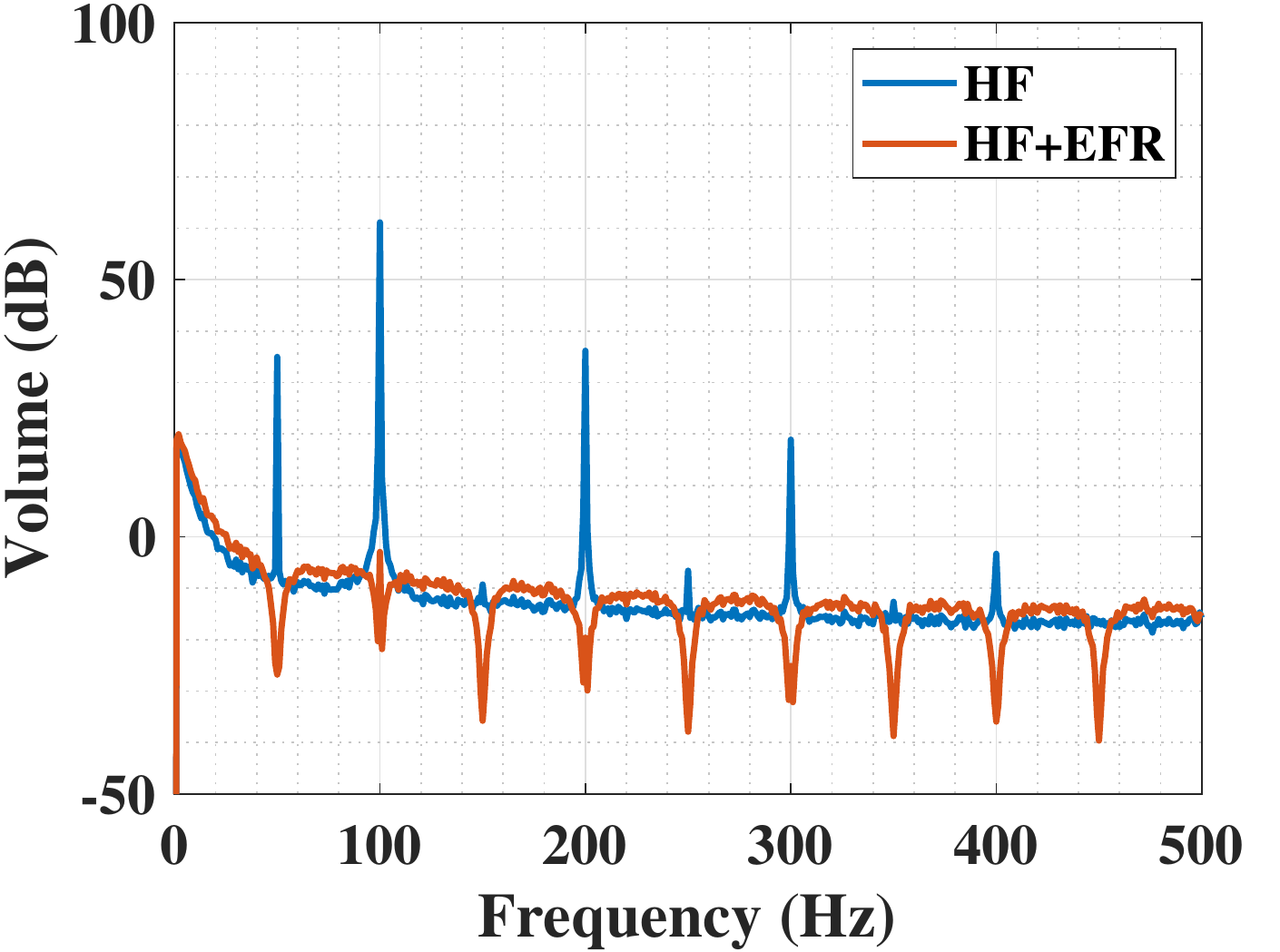} \\
			(c)\small \texttt{Badminton} & (d)\small \texttt{Nerf Gun} \\
		\end{tabular}
	\caption{
		The power spectral density of high-frequency video reconstruction using HF~\cite{Scheerlinck18accv}.
		The 50Hz, 100Hz flicker and their harmonics are effectively attenuated with our event-based flicker removal algorithm.
	}\label{fig:psd}
	\vspace{-2mm}
\end{figure}

\subsection{Main results}
\subsubsection{Video reconstruction and heat map of event rate}
We evaluate the raw and filtered event data using the state-of-the-art pure event-based video reconstruction algorithm E2VID~\cite{Rebecq20pami} in Fig.~\ref{fig:e2vid}.
The flickering light sources are obvious in the E2VID~\cite{Rebecq20pami} reconstructed images using raw event data in Fig.~\ref{fig:e2vid}(a).
After applying our event-based flicker removal (EFR) algorithm, the flashing lights in reconstructed images are clearly attenuated in Fig.~\ref{fig:e2vid}(b).
Please see video materials for better visualisation of the flicker removal performance.

We also provide heat map analysis of event rate to visualise the distribution of the raw event data and the filter event data in Fig.~\ref{fig:e2vid}(c)-(d).
We accumulate event data during a 0.03-second time window for both raw event data and filtered event data.
The event rate is modelled by the ratio of triggered event number to the time period.
The event rate from 0 to the maxima is coloured from blue to yellow in Fig.~\ref{fig:e2vid}. 
In a raw event stream, the number of flicker events is on average 84\% of the total number of events within the image containing a flickering light source. 
%
In Fig.~\ref{fig:e2vid}(c), it is clear that most events are triggered by the flickering fluorescent light, leading to the constantly flickering video reconstructions.
The flicker events contain little or no useful information but occupy most data memory and computational resources, while the actual useful events are triggered at a very low event rate in comparison.
Using our event-based flicker removal algorithm, most of the flicker events are filtered out in Fig.~\ref{fig:e2vid}(d), while preserving useful event data.
Interestingly, objects across the flickering source have minimal impact on our flicker removal algorithm.
For example, when the table tennis ball in Fig.~\ref{fig:e2vid}(b) and the racket in Fig.~\ref{fig:front page}(b) are moving across the flickering fluorescent light, our filter is able to attenuate flickering events, leading to high-quality video reconstruction performance.

We also quantitatively analyse our flicker removal algorithm in Table~\ref{tab:event rate},
evaluating on the same time period as Fig.~\ref{fig:e2vid}.
We use the signal-to-noise ratio (SNR)   \texttt{$\sharp$foreground-events/$\sharp$flicker-events} as a metric for the quality of the data. 
A higher SNR indicates that most of the events in the dataset are relevant to the robot, whereas a lower SNR indicates that many of the events are associated with uninteresting flicker events. 
The first row of Table~\ref{tab:event rate}(a) documents the raw dataset. 
Note that even though the fluorescent light in this scene is less than 5\% of the image, the SNR is an average of 0.19 (or around 84\% of the data is flicker). 
After event flicker removal (Table~\ref{tab:event rate}(b)) the SNR average is 1.07 indicating that at least 50\% of the remaining events are of interest to the robot.
The relative improvement in SNR $(b-a)/(a)$  is shown in the final row of Table~\ref{tab:event rate}. 
On average, the relative improvement in SNR is a factor of 4.63 times better after event flicker removal. 
It should be noted that the SNR is highly dependent on the foreground motion. 
For example, the SNR of the event steam in \texttt{Basketball} dataset is much higher than the others (in all three metrics).
It is because the moving object in the dataset is bigger, triggering more foreground events than other datasets.

\begin{table}[]
	\centering
	\caption{ \label{tab:event rate}
		\em Quantitative analysis of our flicker removal algorithm.
		(a) Signal-to-noise ratio (SNR = \texttt{$\sharp$foreground-events/$\sharp$flicker-events}) of the unfiltered event stream.
		(b) SNR of the filtered event stream.
		(b-a)/(a) Relative improvement in SNR due to event flicker removal.
	}
	\vspace{2mm}
		\begin{tabular}{c|cccc|c}
			\toprule
			\textbf{Metrics}& \texttt{Basketball} & \texttt{Table Tennis} & \texttt{Badminton} & \texttt{Nerf Gun} & \textbf{Avg.}
			\\  \hline \hline
			(a) &  0.24  & 0.17  & 0.19 &   0.17  & 0.19
			\\  \hline
			(b) & 2.08  & 0.92  & 0.72 &  0.56 &  1.07
			\\  \hline \hline
			(b-a)/(a) & 7.67   & 4.41  &  2.79  &  2.29  & 4.63
			\vspace{-1mm}
			\\
			\bottomrule
		\end{tabular}
\end{table}

\begin{figure}[t!]
	\centering
	\begin{tabular}{c}
		\\	\includegraphics[width=0.45\linewidth]{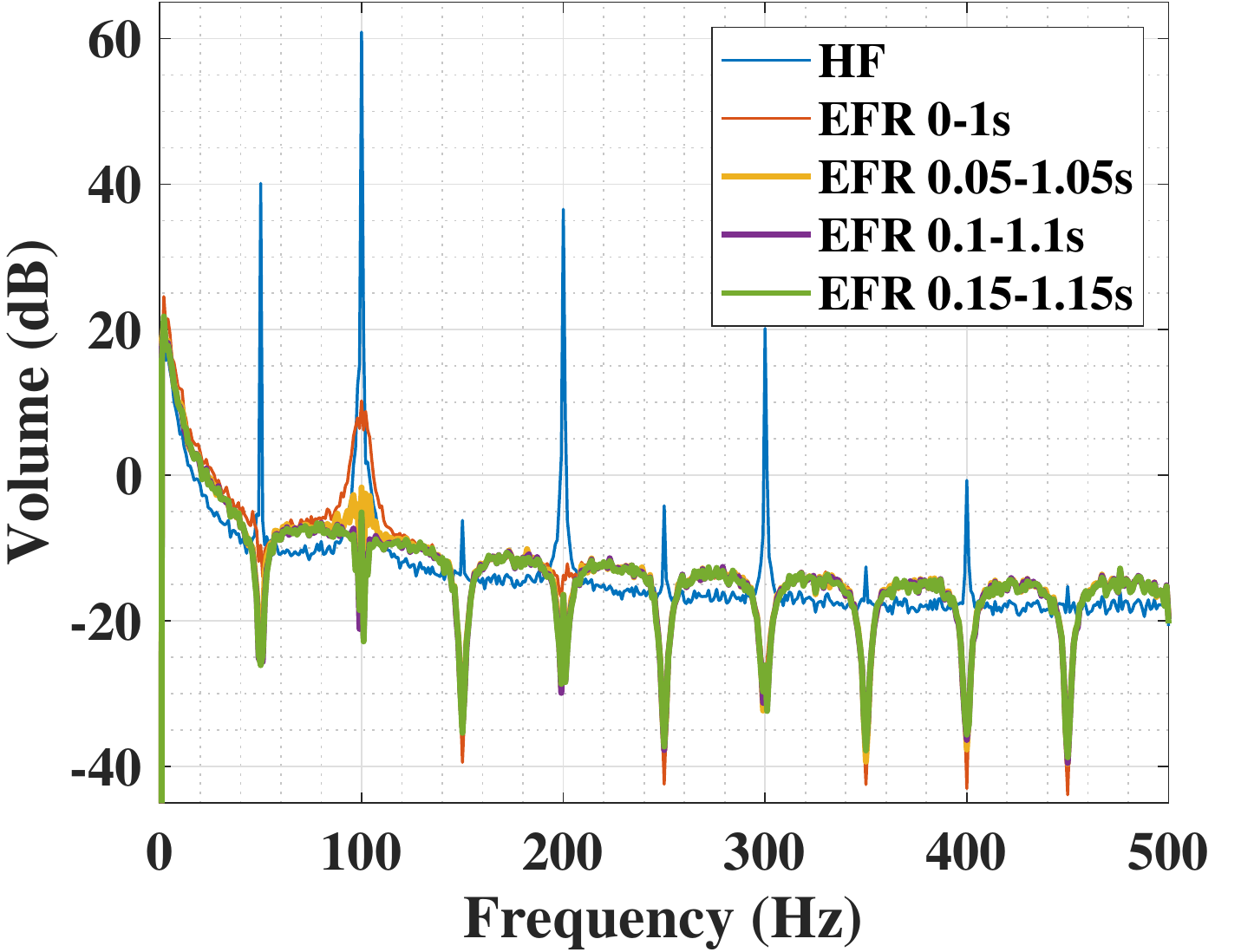}
		\\
	\end{tabular}
	\caption{
		The power spectral density of the high-pass filter and the filtered reconstructed video of the dataset \texttt{Basketball} in the first four 1-second periods.
		In the time period 0-1 seconds, there is still flicker in the reconstructed video, but once the filter state is converged (0.1 seconds later), the most undesired flicker in the video is removed.
	}\label{fig:filter converge}
\end{figure}

\subsubsection{Power spectral density analysis}
The flicker removal performance is also analysed by the power spectral density (PSD) diagrams.
We reconstruct high-frequency video at 1000Hz using a high-pass filter (HF)~\cite{Scheerlinck18accv} and analyse the PSD of a patch at the flickering area before and after applying our flicker removal algorithm.
The PSD diagrams are shown in Fig.~\ref{fig:psd}.
Note that apart from the strong 100Hz and its harmonics, the fluorescent light also triggers events at 50Hz and its harmonics.
Therefore, we set the based frequency to 50Hz to attenuate all flickering frequencies.
Fig.~\ref{fig:psd}(a)(c)(d) are very similar because there is no object moving across the flickering light source.
The table tennis ball in Fig.~\ref{fig:psd}(b) moves across the flickering fluorescent light, leading to more components at low frequencies.
It also affects the flickering frequency of raw event data at 100Hz, and the 100Hz notch of our EFR algorithm is not as deep as the other datasets, but most of the flickering events are still removed.

%
To evaluate the convergence rate of our event-based flicker removal (EFR) algorithm, we also demonstrate the PSD of high-frequency video reconstruction by HF~\cite{Scheerlinck18accv} and flicker filtered reconstruction in the first four 1-second sliding window periods (see Fig.~\ref{fig:filter converge}).
After 0.1 seconds, the PSD plot shows that most frequencies are unchanged while the flickering frequencies have been mostly attenuated.
It shows that our event-based flicker removal algorithm takes around 0.1 seconds to converge.
The actual convergence time also depends on the pre-defined buffer length and the parameter $\rho$.
A smaller $\rho$ will make the converge process even faster, with deeper and wider notches.

\section{CONCLUSIONS}
In this paper, we introduce the problem that event cameras are sensitive to flicker such as from fluorescent or LED lights.
We proposed the first event flicker removal algorithm using a modified linear comb filter and the first event camera de-flickering dataset that includes objects with fast motions under a flickering fluorescent light for evaluation.
Our proposed filter yields a relative improvement 4.63 in the signal-to-noise ratio of the filtered event stream compared to the raw event stream on average, and the flicker removal performance is visualised using power spectral density.
It effectively attenuates flicker events and the associated harmonics,  with convergence within 0.1 seconds.
Our filter can be used as a front-end or preprocessing algorithm to provide a cleaner and flicker-free event stream for other robotics applications.

\clearpage
\bibliographystyle{ieee_fullname}
\bibliography{template_arxiv}

\end{document}